\documentclass{ecai} 

\usepackage{latexsym}
\usepackage{amssymb}
\usepackage{amsmath}
\usepackage{amsthm}
\usepackage{booktabs}
\usepackage{enumitem}
\usepackage{graphicx}
\usepackage{color}
\usepackage{xcolor}
\usepackage{listings}
\newcommand{\BibTeX}{B\kern-.05em{\sc i\kern-.025em b}\kern-.08em\TeX}

\begin{document}


\begin{frontmatter}


\paperid{123} 


\title{ConspEmoLLM: Conspiracy Theory Detection Using an Emotion-Based Large Language Model}


\author[A]{\fnms{Zhiwei}~\snm{Liu}\thanks{Corresponding Author. Email: zhiwei.liu-2@postgrad.manchester.ac.uk}}
\author[A]{\fnms{Boyang}~\snm{Liu}} 
\author[A]{\fnms{Paul}~\snm{Thompson}} 
\author[A]{\fnms{Kailai}~\snm{Yang}}
\author[A,B,C]{\fnms{Sophia}~\snm{Ananiadou}}

\address[A]{The University of Manchester, Manchester, UK}
\address[B]{Artificial Intelligence Research Center, Japan}
\address[C]{Archimedes/Athena RC, Greece}


\begin{abstract}
The internet has brought both benefits and harms to society. A prime example of the latter is misinformation, including conspiracy theories, which flood the web. Recent advances in natural language processing, particularly the emergence of large language models (LLMs), have improved the prospects of accurate misinformation detection.  However, most LLM-based approaches to conspiracy theory detection focus only on binary classification and fail to account for the important relationship between misinformation and affective features (i.e., sentiment and emotions). Driven by a comprehensive analysis of conspiracy text that reveals its distinctive affective features, we propose ConspEmoLLM, the first open-source LLM that integrates affective information and is able to perform diverse tasks relating to conspiracy theories. These tasks include not only conspiracy theory detection, but also classification of theory type and detection of related discussion (e.g., opinions towards theories).  ConspEmoLLM is fine-tuned based on an emotion-oriented LLM using our novel ConDID dataset, which includes five tasks to support LLM instruction tuning and evaluation. We demonstrate that when applied to these tasks, ConspEmoLLM largely outperforms several open-source general domain LLMs and ChatGPT, as well as an LLM that has been fine-tuned using ConDID, but which does not use affective features. ConspEmoLLM can be easily applied to identify and classify conspiracy-related text in the real world. The work has been released at https://github.com/lzw108/ConspEmoLLM/.
\end{abstract}

\end{frontmatter}


\section{Introduction}

Misinformation has become one of the major threats to society. The rise of the internet and social media has made it increasingly simple for misinformation to spread rapidly. Conspiracy theories are one type of misinformation, whose false content is intended to cause harm \cite{napolitano2023conspiracy1}. Examples of popular conspiracy theories include those claiming that the Earth is flat and that vaccines cause autism. Conspiracy theorists ignore scientific evidence and tend to interpret events as secretive actions \cite{giachanou2023detection2}. During the COVID-19 pandemic, the spread of conspiracy theories significantly increased (e.g., the claim that 5G telecommunications networks activated the virus), resulting in a considerable negative impact on society \cite{douglas2021covid3}. As such, there is an increasing urgency for high-performance methods that can automatically detect conspiracy theories.

It has previously been shown that there is a close relationship between misinformation (including conspiracy theories) and affective information, i.e., sentiment and emotions \cite{liu2024emotion3}. For example, \citet{dong2020public4} found that during the COVID-19 pandemic, there was a correlation between the level of public anger and the likelihood of rumor propagation, while \citet{zaeem2020sentiment5} also observed a significant positive correlation between negative emotions and fake news. Based on such observations, many studies have adopted affective information as the means to detect misinformation \cite{liu2023emotion6, zhang2021mining7}. Here, we aim to extend the study of affective information, specifically sentiment and emotions, to deepen our understanding of conspiracy theories and to improve their automated detection.

Pre-trained language models (PLMs) such as BERT \cite{devlin2018bert6} and RoBERTa \cite{liu2019roberta7} have shown outstanding performance when applied to various classification tasks, including conspiracy theory detection \cite{yanagi2021classifying4,peskine2021detecting5}. However, due to their restricted number of parameters, PLMs cannot perform optimally when applied to diverse and complex tasks \cite{zhang2023enhancing8}. Recently, LLMs, which possess significantly larger numbers of parameters, have been explored as a novel means to address the issue of misinformation, with very promising results \cite{hu2023bad9,pavlyshenko2023analysis1,cheung2023factllama2}.  However, these studies mostly focus on binary classification of texts according to whether or not they convey misinformation, Moreover, these previous efforts have mostly utilized simple prompts to test or carry out instruction-tuning of LLMs, or else have employed LLMs as auxiliary tools for other models.   To our knowledge, no existing LLM-based studies have attempted to leverage important affective features that are characteristic of misinformation, nor have such studies attempted to carry out in-depth analyses of conspiracy-related text. 

To address these research gaps, we have constructed a multitask conspiracy detection instruction dataset, \textit{ConDID}, to facilitate instruction-tuning and evaluation of LLMs. Based on annotations in two conspiracy theory datasets, ConDID is divided into five tasks that encompass conspiracy theory judgment, conspiracy theory topic detection, and conspiracy theory intention detection. We subsequently propose a novel open-source LLM, ConspEmoLLM, which is specialized for detecting conspiracy-related information. ConspEmoLLM is created by applying an instruction-tuning method to an emotion-oriented LLM using the ConDID dataset. Evaluation of ConspEmoLLM using the ConDID test set shows that it achieves state-of-the-art (SOTA) performance among other open-source LLMs, as well as the closed-source ChatGPT.

Our main contributions are as follows:

(1) We develop ConDID, the first multi-task conspiracy instruction-tuning dataset.

(2) We conduct an affective analysis of the two conspiracy theory datasets used to construct ConDID, which provides evidence that expressions of sentiment and emotions in conspiracy theory text are distinct from those occurring in the mainstream text.

(3) We propose ConspEmoLLM, the first open-source emotion-based LLM that is specialized for diverse conspiracy theory detection tasks. Evaluation of ConspEmoLLM shows that it outperforms other open-source LLMs and ChatGPT across different tasks.  It also surpasses the performance of 
an instruction-tuned LLM that does not incorporate affective features, thus confirming the effectiveness and importance of affective information in detecting conspiracy-related information.

The remainder of this paper is structured as follows: Section \ref{sec:relatedwork} introduces related work concerning the detection of conspiracy theories and misinformation, analysis of sentiment, and open-source LLMs. Section \ref{sec:method} describes our construction of the ConDID dataset, the affective analysis of conspiracy theory datasets, and the instruction-tuning of ConspEmoLLM. Section \ref{sec:experiment} reports on our evaluation and analyzes of the performance of multiple models on the ConDID test set. Section \ref{sec:conclusion} concludes the paper and provides directions for future work. Sections \ref{sec:limitations} and \ref{sec:ethicsstatement}, respectively, discuss potential limitations and confirm the ethical soundness of our study. The supplementary material provides figures illustrating our detailed affective analysis of conspiracy theory datasets.


\section{Related work \label{sec:relatedwork}}

\subsection{Conspiracy theory and misinformation detection}

PLMs have been widely applied to the task of conspiracy theory and misinformation detection. For example,  \citet{yanagi2021classifying4} utilized BERT as the base model for COVID-19 conspiracy theory detection, while \citet{peskine2021detecting5} applied a domain-specific COVID  BERT (CT-BERT) to the same task. More recently, an increasing amount of attention has been devoted to exploring the application of LLMs to detect conspiracy theories and misinformation. For example, \citet{peskine2023definitions8} employed zero-shot learning to evaluate the accuracy of the GPT-3 model in performing fine-grained multi-label conspiracy theory classification of tweets, while \citet{hu2023bad9} proposed a fake news detection framework that utilizes an LLM as an auxiliary tool to enhance the prediction accuracy of BERT. Meanwhile, \citet{pavlyshenko2023analysis1} employed prompt-based fine-tuning of LLaMA2 for rumor and fake news detection. \citet{cheung2023factllama2} supplemented an LLM with external knowledge to enhance the performance of fake news detection. However, all of these models focus on binary classification and do not exploit affective information for misinformation detection.

\subsection{Affective analysis}

Emotion detection and sentiment analysis are two types of NLP techniques for analyzing human expressions. Sentiment analysis aims to capture both the overall emotional tone conveyed by the data source (typically \textit{positive}, \textit{negative}, or \textit{neutral}) and the intensity of this tone. Emotion detection is the process of categorizing data at a finer level of granularity based on the emotions conveyed. In comparison to sentiments, emotions correspond to more specific and intense feelings. For example,  negative emotions include \textit{anger} and \textit{fear}, while positive emotions include \textit{happiness} and \textit{joy} \cite{liu2024emotion3}. Identifying both sentiments and emotions is crucial for downstream tasks.

Automated affective analysis of text has previously been carried out by various means, including the use of different sentiment analysis tools, such as VADER \cite{hutto2014vader1} and TextBlob\footnote{https://textblob.readthedocs.io/}. PLMs have also been employed for sentiment analysis. For example, \citet{hoang2019aspect3} used BERT for aspect-based sentiment analysis, while \citet{tan2022roberta2} proposed a hybrid sentiment analysis model that combines RoBERTa with an LSTM. More recent work has also begun to explore the effectiveness of LLMs for sentiment analysis. For instance, \citet{zhang2023enhancing8} and \citet{lei2023instructerc4} both utilized retrieval-augmented LLMs to enhance the sentiment analysis capabilities of LLMs when applied to financial news and dialogues, respectively.  Meanwhile, \citet{liu2024emollms} proposed a series of comprehensive LLMs that are specialized for affective analysis (EmoLLMs), and which are capable of analyzing emotions across five different dimensions (i.e. emotion intensity, ordinal classification of emotion intensity, sentiment strength, sentiment classification, emotion detection). EmoLLMs demonstrate strong generalization ability, surpassing ChatGPT and GPT-4 in most emotion analysis tasks. Therefore,  we use one of these EmoLLMs, i.e., EmoLLaMA, to perform affective analysis in this study.

\subsection{Open-source LLMs}

A large amount of research has been dedicated to developing open-source LLMs as an alternative to the well-known closed-source LLMs (e.g., ChatGPT), in order to support easier research into the improvement and application of LLMs.  Popular series of open-source, general language models include LLaMA \cite{touvron2023llama2}, OPT \cite{zhang2022opt6} and BLOOM \cite{workshop2022bloom7}. These are complemented by a range of domain-specific open-source LLMs, including FinMA \cite{xie2023pixiu2} for finance, MentalLLaMA \cite{yang2023mentalllama3} for mental health, ExTES-LLaMA \cite{zheng2023building4} for emotional support chatbots, and TimeLLaMA \cite{yuan2023back1} for temporal reasoning. In this work, we extend the inventory of domain-specific LLMs, by developing the first open-source LLM for multitask conspiracy theory detection based on affective information.



\section{Methods \label{sec:method}}

\subsection{Task formalization}

We approach conspiracy theory detection as a generative task, using a generative model as its foundation. This generative model is an autoregressive language model $P_{\phi}(y|x)$, parameterized using pre-trained weights $\phi$. It differs from previous discriminative models, in terms of its ability to simultaneously handle multiple conspiracy theory detection tasks, i.e.,  conspiracy identification, conspiracy intention detection, and conspiracy theme recognition. Each task, denoted as $t$, is represented as a set of context-target pairs: $D_t={(q_i^t,r_i^t)}_{i={1,2,...N_t}}$, where the context $q$ is a token sequence containing the task description, input text, and query, and $r$ is a further token sequence containing the answer to the query. The model is optimized based on the merged dataset, which combines all task datasets, with the aim of maximizing the objective of conditional language modeling to improve prediction performance.

\subsection{Construction of instruction tuning dataset}

\subsubsection{Raw data}
We build our instruction tuning dataset using two existing annotated datasets: 

\textbf{COCO } The COVID-19 conspiracy theories (COCO) dataset \cite{langguth2023coco2} is an extension of the dataset used in the \textit{MediaEval FakeNews: Corona Virus and Conspiracies Task} challenge  \cite{pogorelov2021fakenews1}.  COCO consists of tweets, each of which is assigned 12 different labels that characterize the \textit{intention} of the tweet with respect to 12 different conspiracy theory categories \footnote{Suppressed Cures, Behavior Control, Anti Vaccination, Fake Virus, Intentional Pandemic, Harmful Radiation, Depopulation, New World Order, Esoteric Misinformation, Satanism, Other Conspiracy Theory, Other Misinformation.}. Each label can have three possible values, i.e., \textit{Unrelated:} The tweet is unrelated to the specific conspiracy category; it contains conspiracy-related keywords, but they are used in a completely different context; \textit{Related:} The tweet is related to the specific category, but it does not propagate misinformation or conspiracy theories; \textit{Conspiracy:} The tweet is related to the specific category and is actively aimed at spreading conspiracy theories.

Each tweet in COCO is also assigned a single overall intention label, as follows: The overall \textit{Conspiracy} label is assigned to tweets for which the \textit{Conspiracy} label is used for at least one of the 12 categories. Otherwise, if the \textit{Related} category is assigned to at least one of the categories, then the overall label of \textit{Related} is used.  The overall label of \textit{Unrelated} is only used for tweets that are unrelated to all 12 conspiracy categories.  

\textbf {LOCOAnnotations }  The Language of Conspiracy (LOCO) \cite{miani2021loco3} is a corpus consisting of documents gathered from the internet (88 million words), which is used to study the differences between conspiracy language and mainstream language. We use an annotated subset of LOCO, created by \citet{mompelat2022loco4}, which we refer to as \textit{LOCOAnnotations}. This subset consists of documents concerning two different topics (i.e., the Sandy Hook school shooting and coronavirus) in which two types of labels have been assigned. The first type of label concerns whether or not the document is directly concerned with a conspiracy theory. The second type of label reflects the degree of relatedness to a conspiracy theory, using three labels (\textit{closely related, broadly related} or \textit{not related}). 

\subsubsection{Tasks}

Using these two different datasets, we define five different tasks. Tasks 1-3 are based on the COCO corpus, and are similar to those used in the MediaEval FakeNews challenge, while Tasks 4 and 5 are based on LOCOAnnotations.

\textbf{Task 1: Conspiracy Intention Detection } Determine the overall intention of the tweet (i.e., \textit{Unrelated/Related/Conspiracy}) towards COVID-19 conspiracy theories. 

\textbf{Task 2: Conspiracy Theory Topics Detection } Determine whether the tweet mentions or refers to any of the 12 predefined conspiracy theory categories.

\textbf{Task 3: Combination of Task 1 and Task 2 } Predict \textit{both} the conspiracy category and the relationship of the tweet (\textit{Unrelated/Related/Conspiracy}) to the category.

\textbf{Task 4: Conspiracy Theory Detection } Determine whether a document is directly concerned with a conspiracy theory  (\textit{Conspiracy/Non-Conspiracy}).

\textbf{Task 5: Relatedness Detection } Determine the level of relatedness of a document to a conspiracy theory (\textit{Closely related/Broadly related/Not related}).

\subsubsection{Affective analysis of raw data}

EmoLLMs \cite{liu2024emollms} is a series of affective analysis models. We adopt the best-performing model from this series (i.e., EmoLLaMA-chat-7b) to conduct sentiment analysis on COCO and LOCOAnnotations across five different dimensions, as follows:
\begin{itemize}
\item  \textit{Emotion intensity score}: For each of four different emotions (\textit{anger, fear, joy} and  \textit{sadness}), a real-valued score between 0 and 1 is assigned to represent the intensity of the emotion in the text.
\item \textit{Emotion intensity classification}: For the same four emotions, one of four different emotional intensity classes is assigned to the text,  i.e. \textit{\{no, low, moderate, high\} emotional intensity}.
\item \textit{Sentiment strength score}: A real-valued score between 0 (most negative) and 1 (most positive) is assigned to represent the intensity of sentiment expressed in the text.
\item \textit{Sentiment classification}: One of seven classes (i.e. \textit{\{very, moderately, slightly\} negative, neutral, \{slightly, moderately, very\} positive}) is assigned to represent the intensity of sentiment conveyed in the text.
\item \textit{Emotion detection}: One or more labels is assigned to the text if any of eleven different emotions are conveyed (i.e. \textit{anger, anticipation, disgust, fear, joy, love, optimism, pessimism, sadness, surprise} and  \textit{trust}). If none of these emotions is expressed, then the label \textit{neutral or no emotion} is assigned.
\end{itemize}

\begin{figure}[htb]
\centering
\includegraphics[width=\columnwidth]{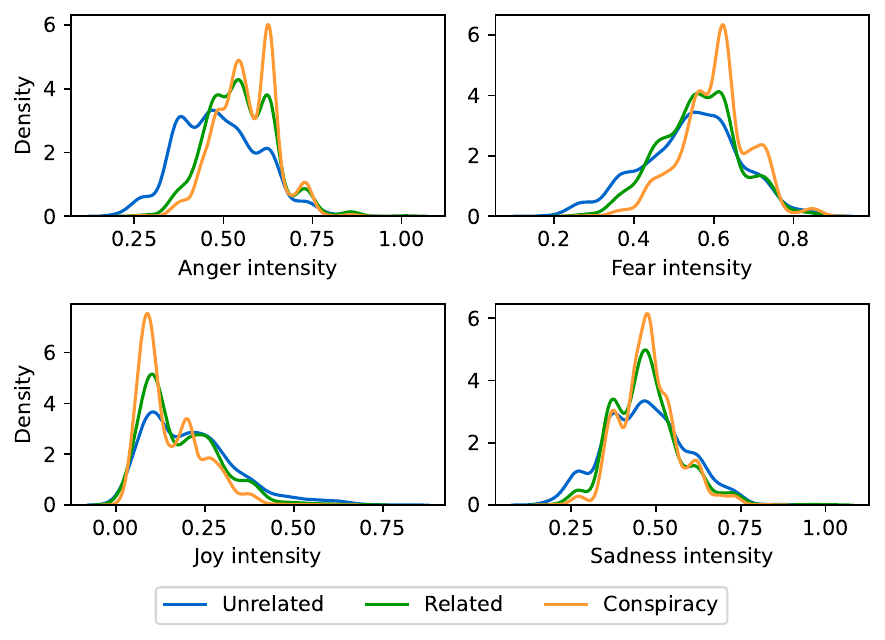}
\caption{Emotion intensity of different intentions}
\label{fig:COCO_EI_label}
\end{figure}

\begin{figure}[htb]
\centering
\includegraphics[width=\columnwidth]{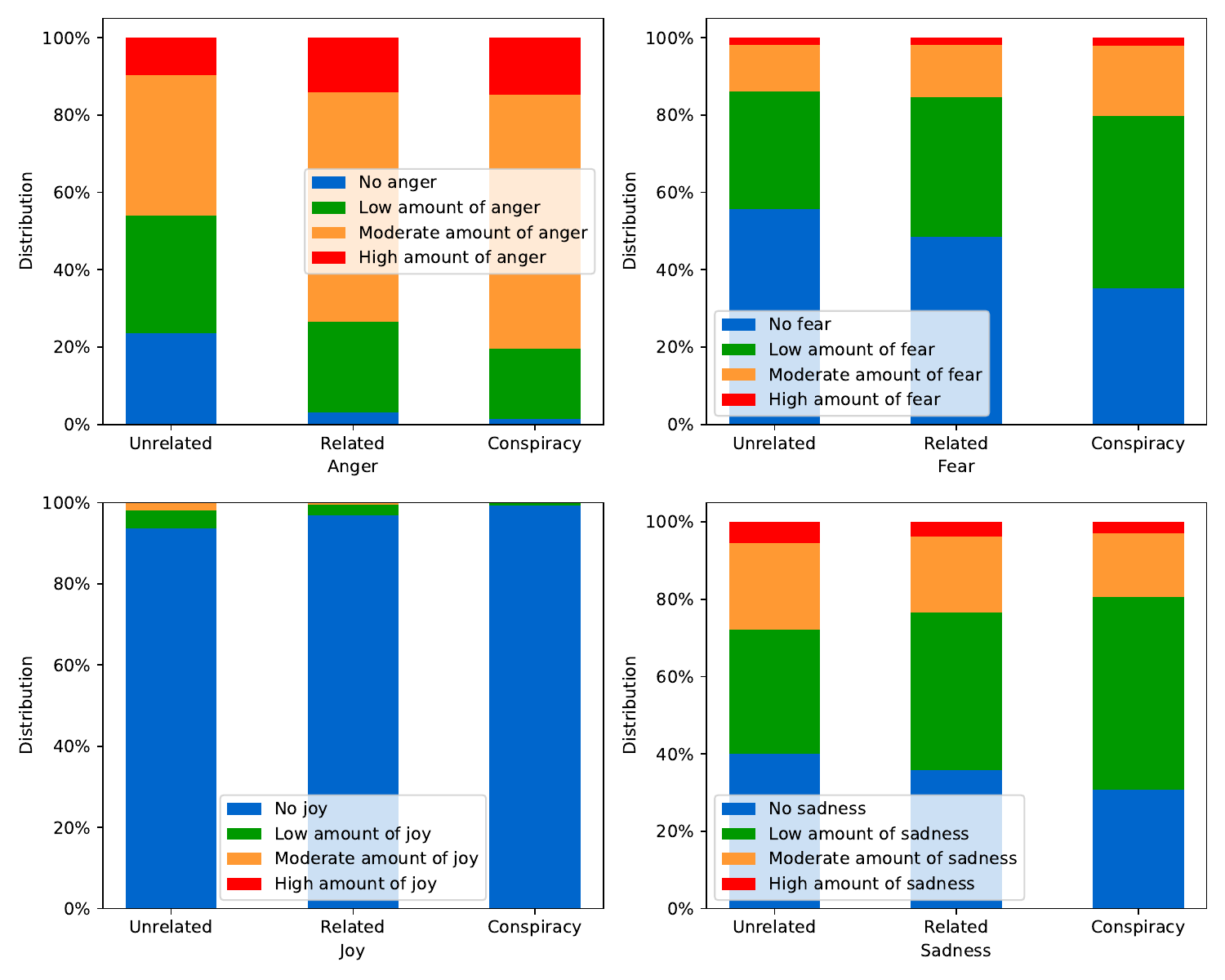}
\caption{Emotion intensity classification of different intentions}
\label{fig:COCO_EIoc_label}
\end{figure}

\begin{figure}[htb]
\centering
\includegraphics[width=\columnwidth]{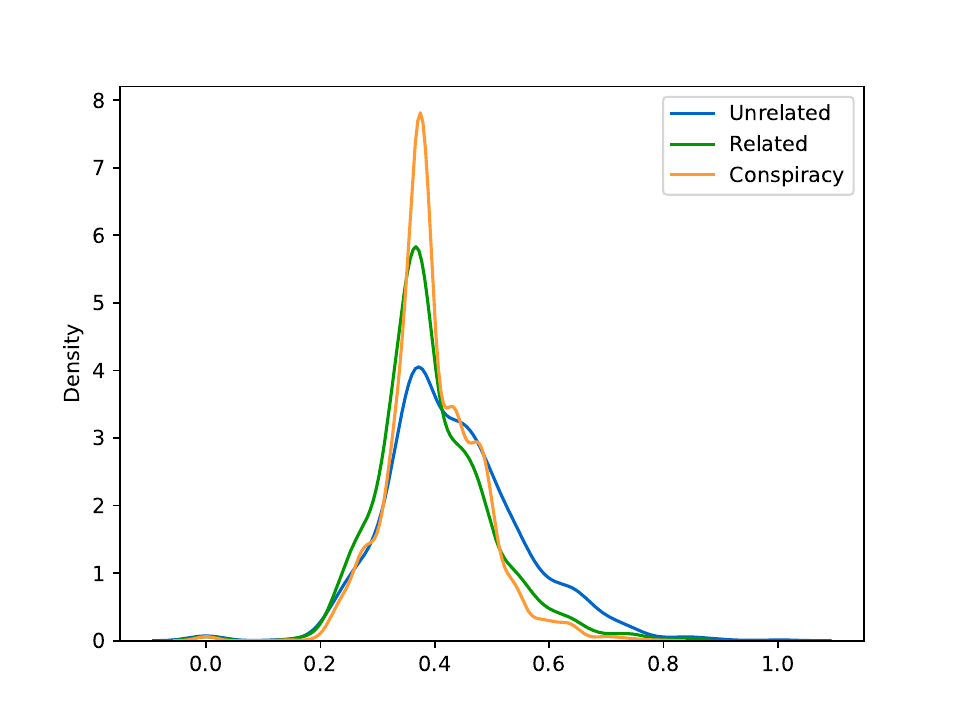}
\caption{Sentiment strength of different intentions}
\label{fig:COCO_Vreg_label}
\end{figure}

\begin{figure}[htb]
\centering
\includegraphics[width=\columnwidth]{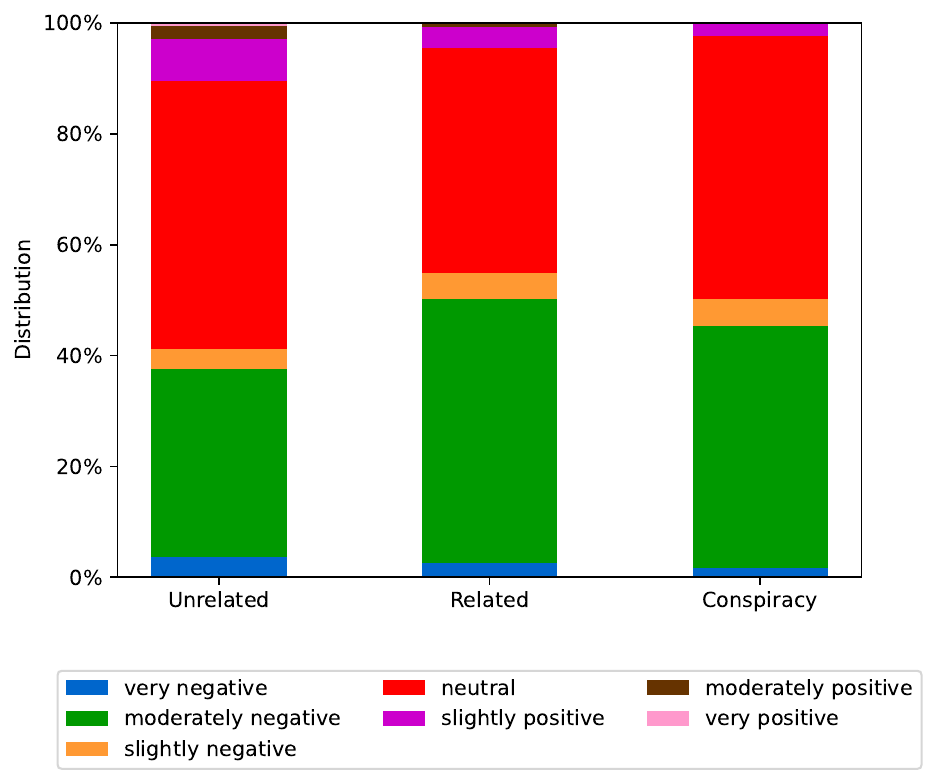}
\caption{Sentiment classification of different intentions}
\label{fig:COCO_Voc_label}
\end{figure}

\begin{figure}[htb]
\centering
\includegraphics[width=0.9\columnwidth]{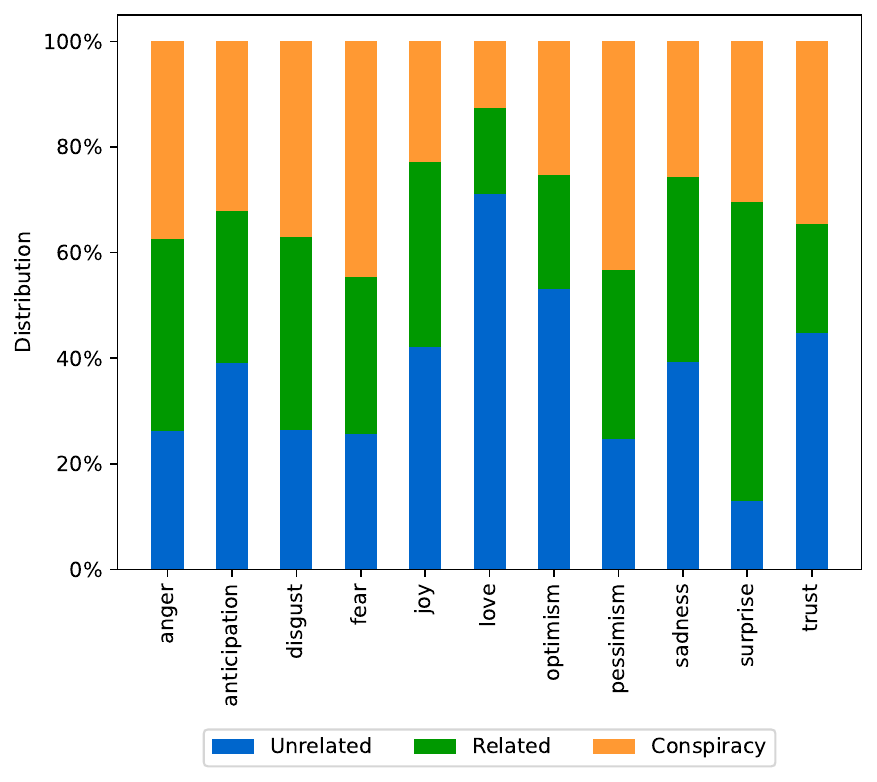}
\caption{Emotion classification of different intentions}
\label{fig:COCO_Ec_label}
\end{figure}

The results of analyzing the COCO dataset according to these five dimensions are shown in Figures \ref{fig:COCO_EI_label} to \ref{fig:COCO_Ec_label}, broken down according to the different intentions of the tweets in the dataset (i.e., \textit{Unrelated/Related/Conspiracy}). In Figures \ref{fig:COCO_EI_label} and \ref{fig:COCO_Vreg_label}, the x-axis represents the real-valued scores for intensity of different emotions and sentiment strength, respectively. The y-axis represents the corresponding probability density distribution (i.e., the number of tweets belonging to each intention class that has a particular score for emotional intensity/sentiment strength). In Figures \ref{fig:COCO_EIoc_label}, \ref{fig:COCO_Voc_label} and \ref{fig:COCO_Ec_label}, the y-axis represents the distribution of labels within the intention class indicated on the x-axis. Due to space limitations, these figures only report on part of our complete analysis. The supplementary material includes additional figures that depict an analysis of the five dimensions based on the different conspiracy categories of tweets in the COCO dataset. The supplementary material also includes analyses of LOCOAnnoations, which examine affective differences between conspiracy and non-conspiracy text, and between text exhibiting various levels of relatedness to conspiracy theories (i.e., \textit{Not related/closely related/broadly related}).


Figure \ref{fig:COCO_EI_label} presents a quantitative analysis of emotional intensity. It may be observed that tweets that refer to conspiracy theories (i.e., \textit{Related/Conspiracy}) express stronger feelings of anger and fear, with a lesser degree of joy, compared to the \textit{Unrelated} tweets. The intensity of sadness is similar across the three categories, indicating that during the pandemic, people were generally in a state of sadness. Similar conclusions may be drawn from Figure \ref{fig:COCO_EIoc_label}, which presents a qualitative analysis of emotional intensity. The sentiment strength in Figure \ref{fig:COCO_Vreg_label} and the sentiment polarity classification in Figure \ref{fig:COCO_Voc_label} indicate that sentiments expressed in \textit{Unrelated} tweets are more strongly positive than sentiments in tweets related to conspiracy theories. It can be observed from Figure \ref{fig:COCO_Ec_label} that tweets related to conspiracy theories predominantly convey negative sentiments and emotions (e.g., anger, fear, and disgust). In contrast, the \textit{Unrelated} tweets are more likely to express positive sentiments and emotions (e.g., joy, love, and optimism). The figures in the supplementary materials reveal that the types of emotions expressed in tweets can also vary according to the specific category of conspiracy theory being discussed in the COCO dataset, and that there are noticeable differences in the types of emotion information conveyed across different categories and levels of conspiracy relatedness in the LOCOAnnotations dataset. Overall, it can be inferred that there are close relationships between affective information and conspiracy theories. The affective information can be an important feature for detecting conspiracy theories.

\subsubsection{Construction of the ConDID conspiracy detection instruction dataset}

\begin{table}[!t]
\centering
\begin{tabular}{lp{2.5cm}p{3cm}}
\hline
Task   & Raw (Train/Dev/Test)          & Instruction (Train/Dev/Test) \\ \hline
Task 1 &                               & 2092/697/698                 \\
Task 2 & 2092/697/698                  & 2092/697/698                 \\
Task 3 &                               & 25104/8364/8376              \\
Task 4 &                               & 669/223/233                  \\
Task 5 & 669/223/233                   & 669/223/233                  \\ \hline
\end{tabular}
\caption{\label{tab:datastatistics}
Dataset statistics. \textit{Raw} denotes the raw data from COCO and LOCOAnnotations. \textit{Instruction} denotes the converted instruction data based on \textit{Raw}.}
\end{table}

\begin{table*}[!t]
\begin{tabular}{p{1cm}p{16cm}}
\hline
Task   & Prompt Template                                                                                                                                                                                                                                                                                                                                                                                                    \\ \hline
Task 1 & Task:  Classify the text regarding COVID-19   conspiracy theories or misinformation into one of the following three classes: 0. Unrelated. 1. Related (but not supporting). 2. Conspiracy   (related and supporting).                                                                                                                                                                                                         \\
Task 2 & Task: detect whether the text in any form mentions or refers to any of the specific categories of COVID-19   conspiracy theories  ('suppressed cures', 'behavior control', 'anti vaccination', 'fake virus', 'intentional pandemic',   'harmful radiation', 'depopulation', 'new world order', 'satanism', 'esoteric misinformation',  'other conspiracy theory') or other misinformation. If it doesn't, it is 'no conspiracy. \\
Task 3 & Task: Classify the text regarding the specific category [Specific Conspiracy] into one of the following three classes: 0. Unrelated. 1. Related (but not supporting). 2. Conspiracy (related and supporting).                                                                                                                                                                                                                   \\
Task 4 & Task: Determine if the text is a conspiracy theory. Classify it into one of the following two classes: 0. non-conspiracy. 1. conspiracy.                                                                                                                                                                                                                                                                                        \\
Task 5 & Task: Determine the relatedness between the text and [conspiracy theory]. Classify it into one of the following three classes: 0. not related. 1. closely related. 2. broadly related.                                                                                                                                                                                                                                          \\ 
Affective prompt & Task: original task prompt +   "You can also refer to the affective information. (1) Emotion intensity:   anger: 0.521, fear: 0.625, joy: 0.25, sadness: 0.354. (2) Ordinal   classification of emotion intensity: moderate amount of anger can be   inferred. low amount of fear can be inferred. no joy can be inferred. no   sadness can be inferred. (3) Sentiment intensity: 0.435. (4) Sentiment   classification: neutral or mixed mental state can be inferred. (5) The   emotions included are: anger, disgust, fear."
 \\
 \hline
\end{tabular}
\caption{\label{tab:taskprompt} Prompts used for each task.}
\end{table*}

\begin{figure*}[htb]
\centering
\includegraphics[width=1.9\columnwidth]{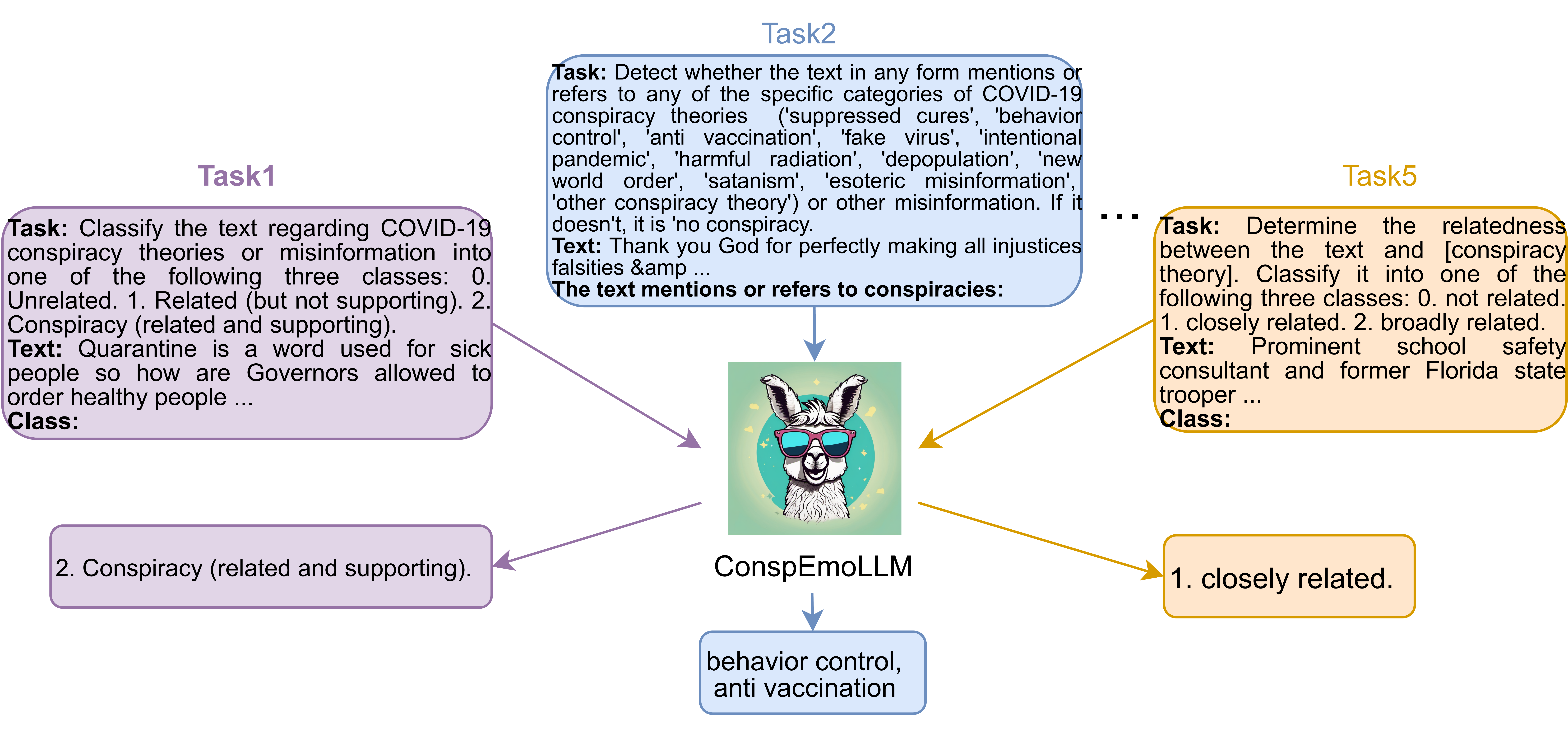}
\caption{An overview of multi-task instruction tuning of ConspEmoLLM}
\label{fig:ConspEmoLLM}
\end{figure*}

We used the raw datasets as the basis to build the instruction dataset. We randomly selected 20\% of the data as the test set and 20\% as the validation set. The dataset statistics are presented in Table 1. For Task 3, prompts needed to be created for each of the 12 categories, resulting in a magnitude of data that is 12 times the size of the original corpus. We constructed instruction-tuning data for each task based on the following template: 

\begin{center}
\fcolorbox{black}{lightgray}{
\begin{minipage}{0.45\textwidth}
Task: \textit{[task prompt]}  Text: \textit{[input text]}  Class/The text mentions or refers to conspiracies: \textit{[output]}
\end{minipage}
}
\end{center}

\textit{[task prompt]} denotes the instruction for the task. \textit{[input text]} is a data item from the raw data. With the exception of Task 2, all tasks use \textit{Class} to prompt LLM to generate answers. \textit{[output]} is the output from LLM. Table \ref{tab:taskprompt} lists the task prompts for each task, and Figure \ref{fig:ConspEmoLLM} presents several examples used to fine-tune the LLM. To allow an evaluation of the impact of explicitly encouraging the LLMs to make use of affective information, we additionally constructed prompts of the form shown in the final row of Table \ref{tab:taskprompt}. These prompts provide information about the specific analysis results obtained from EmoLLaMA regarding the text to be classified.

\subsection{ConspEmoLLM and ConspLLM}

We built ConspEmoLLM by fine-tuning EmoLLaMA-chat-7b \cite{liu2024emollms} using the ConDID dataset. We also fine-tuned an LLM that does not use affective information (ConspLLM) based on LLaMA2-chat-7b \cite{touvron2023llama2}, also using ConDID. The models are trained based on the AdamW optimizer \cite{loshchilov2017decoupled4} for three epochs, using DeepSpeed \cite{rasley2020deepspeed1} to reduce memory usage. We set the batch size to 256. The initial learning rate is set to 1e-6 with a warm-up ratio of 5\%, and the maximum model input length is set to 4096. All models are trained on two Nvidia Tesla A100 GPUs, each with 80GB of memory. Figure \ref{fig:ConspEmoLLM} provides an overview of multi-task instruction tuning of ConspEmoLLM for diverse conspiracy detection tasks. 


\begin{table*}[h]
\centering
\begin{tabular}{lp{0.6cm}p{0.6cm}p{0.6cm}p{0.6cm}p{0.6cm}p{0.6cm}p{0.6cm}p{0.6cm}p{0.6cm}p{0.6cm}p{0.6cm}p{0.6cm}}
\hline
Model        & \multicolumn{4}{c}{Task1}                                         & \multicolumn{4}{c}{Task2}                                         & \multicolumn{4}{c}{Task3}                                         \\
             & ACC            & PRE            & REC            & F1             & ACC            & PRE            & REC            & F1             & ACC            & PRE            & REC            & F1             \\ \hline
BERT         & 0.576          & 0.601          & 0.428          & 0.406          & 0.272          & 0.150          & 0.012          & 0.023          & 0.893          & 0.871          & 0.893          & 0.842          \\
RoBERTa      & 0.517          & 0.505          & 0.335          & 0.231          & 0.279          & 0.204          & 0.077          & 0.112          & 0.893          & 0.842          & 0.893          & 0.844          \\
CT-BERT      & 0.564          & 0.472          & 0.469          & 0.428          & 0.042          & 0.137          & 0.303          & 0.175          & 0.893          & 0.797          & 0.893          & 0.842          \\ \hline
Falcon       & 0.265          & 0.361          & 0.265          & 0.211          & 0.189          & 0.193          & 0.180          & 0.160          & 0.556          & 0.810          & 0.556          & 0.653          \\
Vicuna       & 0.380          & 0.446          & 0.380          & 0.392          & 0.136          & 0.172          & 0.270          & 0.150          & 0.437          & 0.825          & 0.437          & 0.556          \\
LLaMA2-chat  & 0.325          & 0.527          & 0.325          & 0.251          & 0.193          & 0.093          & 0.071          & 0.074          & 0.674          & 0.814          & 0.674          & 0.732          \\
OPT          & 0.311          & 0.389          & 0.311          & 0.298          & 0.235          & 0.216          & 0.048          & 0.072          & 0.481          & 0.805          & 0.481          & 0.590          \\
BLOOM        & 0.328          & 0.389          & 0.328          & 0.318          & 0.268          & 0.000          & 0.000          & 0.000          & 0.114          & 0.810          & 0.114          & 0.146          \\
ChatGPT      & 0.638          & 0.677          & 0.638          & 0.596          & 0.324          & 0.546          & 0.333          & 0.332          & 0.208          & 0.896          & 0.208          & 0.240          \\
ChatGPT-aff  & 0.583          & 0.647          & 0.583          & 0.525          & 0.312          & 0.449          & 0.326          & 0.308          & 0.150          & 0.898          & 0.150          & 0.146          \\ \hline
ConspLLM     & 0.662          & \textbf{0.757} & 0.662          & 0.675          & 0.328          & 0.685          & 0.320          & 0.334          & 0.893          & 0.886          & 0.893          & \textbf{0.864} \\
ConspLLM-aff & 0.517          & 0.483          & 0.517          & 0.354          & 0.203          & 0.404          & 0.330          & 0.301          & 0.077          & \textbf{0.901} & 0.077          & 0.015          \\
ConspEmoLLM  & \textbf{0.695} & 0.755          & \textbf{0.695} & \textbf{0.705} & \textbf{0.340} & \textbf{0.699} & \textbf{0.345} & \textbf{0.364} & \textbf{0.897} & 0.884          & \textbf{0.897} & 0.860          \\ \hline
\end{tabular}
\caption{\label{tab:resultsoncoco}
Evaluation results for Tasks 1, 2, and 3}
\end{table*}

\begin{table*}[h]
\centering
\begin{tabular}{lcccccccc}
\hline
Model        & \multicolumn{4}{c}{Task4}                                         & \multicolumn{4}{c}{Task5}                                         \\
             & ACC            & PRE            & REC            & F1             & ACC            & PRE            & REC            & F1             \\ \hline
BERT         & 0.691          & 0.677          & 0.642          & 0.645          & \textbf{0.614} & 0.623          & 0.356          & 0.296          \\
RoBERTa      & 0.677          & 0.664          & 0.670          & 0.666          & 0.601          & 0.200          & 0.333          & 0.250          \\ \hline
Falcon       & 0.448          & 0.489          & 0.448          & 0.453          & 0.422          & 0.640          & 0.422          & 0.477          \\
Vicuna       & 0.529          & 0.524          & 0.529          & 0.526          & 0.274          & 0.633          & 0.274          & 0.279          \\
LLaMA2-chat  & 0.583          & 0.611          & 0.583          & 0.588          & 0.291          & 0.634          & 0.291          & 0.314          \\
OPT          & 0.466          & 0.509          & 0.466          & 0.470          & 0.507          & \textbf{0.676} & 0.507          & 0.554          \\
BLOOM        & 0.439          & 0.483          & 0.439          & 0.443          & 0.587          & 0.653          & 0.587          & 0.617          \\
ChatGPT      & 0.668          & \textbf{0.774} & 0.668          & 0.664          & 0.596          & 0.576          & 0.596          & 0.574          \\
ChatGPT-aff  & 0.673          & 0.769          & 0.673          & 0.670          & 0.587          & 0.557          & 0.587          & 0.551          \\ \hline
ConspLLM     & 0.641          & 0.663          & 0.641          & 0.646          & 0.596          & 0.574          & 0.596          & 0.580          \\
ConspLLM-aff & \textbf{0.731} & 0.758          & \textbf{0.731} & \textbf{0.735} & 0.453          & 0.555          & 0.453          & 0.479          \\
ConspEmoLLM  & 0.700          & 0.717          & 0.700          & 0.703          & 0.610          & 0.647          & \textbf{0.610} & \textbf{0.623} \\ \hline
\end{tabular}
\caption{\label{tab:resultsonloco}
Evaluation results for Tasks 4 and 5}
\end{table*}

\section{Experiments \label{sec:experiment}}

\subsection{Baseline models}

\textbf{PLMs:} Conspiracy theory detection is typically regarded as a classification task. For our baseline models, we selected commonly used PLMs, which can only be fine-tuned for individual tasks, i.e., the general language BERT and RoBERTa, along with CT-BERT \cite{muller2023covid1}, which is tailored for the COVID-19 domain. We treat Tasks 1 and 5 as 3-way classification tasks, and Task 4 as a binary classification task, using cross-entropy loss for training. Task 2 is treated as a multi-label binary classification problem, which can be achieved by utilizing the binary cross-entropy with logits loss. To address Task 3, fine-tuning is performed for 12 different classification problems, each with its own cross-entropy loss function. The final loss is the average of the 12 losses.

\textbf{LLMs: } LLMs have been proven to be capable of solving numerous tasks. We apply zero-shot prompting on the instruction dataset to the following open-source LLMs:  Falcon-7B-instruct \cite{penedo2023refinedweb}, LLaMA2-chat-7B \cite{touvron2023llama2}, OPT-7B \cite{zhang2022opt6}, BLOOM-7B \cite{workshop2022bloom7}, and Vicuna-7B-v1.5\footnote{https://huggingface.co/lmsys/vicuna-13b-v1.5}. We also utilize zero-shot prompting with the proprietary LLM ChatGPT.

\subsection{Evaluation methods}

Since the tasks addressed in this paper are all classification problems, we apply commonly used metrics to evaluate the performance of the models, i.e., Accuracy (ACC), Precision (PRE), Recall (REC), and weighted F1 score.

\subsection{Results}

Tables \ref{tab:resultsoncoco} and \ref{tab:resultsonloco} report the results for each task. ChatGPT-aff and ConspLLM-aff are the models that use the prompts containing explicit affective information. From the results tables, we can observe that, in terms of F1 score,  the fine-tuned ConspLLM and ConspEmoLLM outperform all other open-source models\footnote{It should be noted that the LLMs that are not instruction-tuned produced some responses that do not follow instructions, i.e., they do not produce output of the type requested in the prompts. In such cases, we label them as \textit{unrelated} or \textit{non-conspiracy}, depending on the task.}, as well the PLMs.  ConspLLM and ConspEmoLLM also both outperform ChatGPT, with the exception of ConspLLM on Task 4.  Moreover, ConspEmoLLM, which is fine-tuned based on a large emotion language model, achieves F1 scores that are over 3\% higher than ConspLLM for all tasks except for Task 3, in which it lags slightly behind ConspLLM. The results for ChatGPT-aff and ConspLLM-aff reveal that explicitly augmenting prompts with affective information leads to reduced performance for all tasks apart from Task 4.  We can infer that explicitly adding affective information appears to distract models' attention from the task at hand. In contrast, the implicit use of emotion information by ConspEmoLLM is more successful in allowing emotional cues to be leveraged.

\section{Automatic predictions \label{sec:predictions}}

The following example code demonstrates how ConspEmoLLM can be applied to automatically classify text according to whether or not it represents a conspiracy theory. Any piece of text may be classified by ConspEmoLLM, by replacing \textit{[input text]} with the text of interest. The task prompt may also be adjusted to perform the other conspiracy-related tasks introduced above. Further details can be found on Github \footnote{https://github.com/lzw108/ConspEmoLLM}.

\hspace*{\fill}

\begin{lstlisting}
from transformers import AutoTokenizer, AutoModelForCausalLM
Model_PATH = "lzw1008/ConspEmoLLM-7b"
tokenizer = AutoTokenizer.from_pretrained(MODEL_PATH)
model = AutoModelForCausalLM.from_pretrained(MODEL_PATH, device_map='auto')

prompt = '''Human: 
Task: Task: Determine if the text is a conspiracy theory. Classify it into one of the following two classes: 0. non-conspiracy. 1. conspiracy
Text: [input text]
Class:
Assistant:
'''
inputs = tokenizer(prompt, return_tensors="pt")
generate_ids = model.generate(inputs["input_ids"], max_length=256)
response = tokenizer.batch_decode(generate_ids, skip_special_tokens=True)[0]
print(response)
>> 0. non-conspiracy / 1. conspiracy
\end{lstlisting}

\section{Conclusion \label{sec:conclusion}}

In this paper, our comprehensive affective analysis of two conspiracy theory datasets has demonstrated that conspiracy theory text exhibits sentiment and emotion features that are distinct from mainstream text.  The results of this analysis motivated our development of ConspEmoLLM, an open-source domain-specific LLM that is based on affective information, and which can perform diverse conspiracy theory detection tasks. ConspEmoLLM was fine-tuned using our newly constructed multitask conspiracy detection instruction dataset (ConDID).  Evaluation on the test set of ConDID reveals that ConspEmoLLM achieves SOTA performance among the other open-source LLMs tested, as well as ChatGPT, in all tasks. On most tasks, the performance of ConspEmoLLM surpasses that of the ConspLLM model, which was also instruction-tuned using ConDID, but does not use affective information. These results provide strong evidence of the importance of affective features in detecting various types of information relating to conspiracy theories. We also exemplified how ConspEmoLLM can be applied straightforwardly to detect the presence of conspiracy-related information in any piece of text. This demonstrates the real world utility of our model in helping people to identify the misinformation on the Internet, thus contributing towards reducing the potential harm caused by conspiracy theories.

As future work, we aim to augment the ConDID dataset with further conspiracy theory datasets, including data from multiple platforms, sources, domains and languages. This should help to further improve the performance and diversity of tasks that can be carried out using ConspEmoLLM. We will additionally explore alternative methods of incorporating affective information to further improve the ability of the model to detect conspiracy theories. Furthermore, we will design more appropriate prompts and utilize more complex and diverse model structures to better leverage emotions and sentiments.

\section{Limitations \label{sec:limitations}}

The potential limitations of our work may be summarized as follows: 

(1) Due to restricted computational resources, we only carried out instruction-tuning and evaluation of conspiracy theory detection tasks using 7B LLMs. As such, we have not considered how the use of larger or different model architectures may potentially impact upon performance in conspiracy theory detection tasks.

(2) The datasets used in this paper mostly concern COVID-19 conspiracy theories and are limited in size. These limitations may affect the ability of the model to generalize to other types of data or domains. However,  as mentioned in Section \ref{sec:conclusion}, we plan to increase the size of our ConDID dataset, by collecting additional data from diverse sources and domains. It is hoped that this will help to improve both the performance and generalizability of the model. 

\section{Ethics Statement \label{sec:ethicsstatement}}

The original datasets collected to construct the ConDID dataset are sourced from public social media platforms and websites. We strictly adhere to privacy agreements and ethical principles to protect user privacy and to ensure the proper application of anonymity in all texts.



\begin{ack}
The ConspEmoLLM illustration in Figure \ref{fig:ConspEmoLLM} was generated using PIXLR\footnote{https://pixlr.com/image-generator/}. 
This work is supported by the computational shared facility at the University of Manchester and the scholar award from the Department of Computer Science at the University of Manchester. This work is also supported by the Centre for Digital Trust and Society at the University of Manchester, the Manchester-Melbourne-Toronto Research Fund, and the New Energy and Industrial Technology Development Organization.
\end{ack}



\bibliography{mP54}

\begin{thebibliography}{37}
\providecommand{\natexlab}[1]{#1}
\providecommand{\url}[1]{\texttt{#1}}
\expandafter\ifx\csname urlstyle\endcsname\relax
  \providecommand{\doi}[1]{doi: #1}\else
  \providecommand{\doi}{doi: \begingroup \urlstyle{rm}\Url}\fi

\bibitem[Cheung and Lam(2023)]{cheung2023factllama2}
T.-H. Cheung and K.-M. Lam.
\newblock Factllama: Optimizing instruction-following language models with external knowledge for automated fact-checking.
\newblock In \emph{2023 Asia Pacific Signal and Information Processing Association Annual Summit and Conference (APSIPA ASC)}, pages 846--853. IEEE, 2023.

\bibitem[Devlin et~al.(2018)Devlin, Chang, Lee, and Toutanova]{devlin2018bert6}
J.~Devlin, M.-W. Chang, K.~Lee, and K.~Toutanova.
\newblock Bert: Pre-training of deep bidirectional transformers for language understanding.
\newblock \emph{arXiv preprint arXiv:1810.04805}, 2018.

\bibitem[Dong et~al.(2020)Dong, Tao, Xia, Ye, Xu, Jiang, and Liu]{dong2020public4}
W.~Dong, J.~Tao, X.~Xia, L.~Ye, H.~Xu, P.~Jiang, and Y.~Liu.
\newblock Public emotions and rumors spread during the covid-19 epidemic in china: web-based correlation study.
\newblock \emph{Journal of Medical Internet Research}, 22\penalty0 (11):\penalty0 e21933, 2020.

\bibitem[Douglas(2021)]{douglas2021covid3}
K.~M. Douglas.
\newblock Covid-19 conspiracy theories.
\newblock \emph{Group Processes \& Intergroup Relations}, 24\penalty0 (2):\penalty0 270--275, 2021.

\bibitem[Giachanou et~al.(2023)Giachanou, Ghanem, and Rosso]{giachanou2023detection2}
A.~Giachanou, B.~Ghanem, and P.~Rosso.
\newblock Detection of conspiracy propagators using psycho-linguistic characteristics.
\newblock \emph{Journal of Information Science}, 49\penalty0 (1):\penalty0 3--17, 2023.

\bibitem[Hoang et~al.(2019)Hoang, Bihorac, and Rouces]{hoang2019aspect3}
M.~Hoang, O.~A. Bihorac, and J.~Rouces.
\newblock Aspect-based sentiment analysis using bert.
\newblock In \emph{Proceedings of the 22nd nordic conference on computational linguistics}, pages 187--196, 2019.

\bibitem[Hu et~al.(2023)Hu, Sheng, Cao, Shi, Li, Wang, and Qi]{hu2023bad9}
B.~Hu, Q.~Sheng, J.~Cao, Y.~Shi, Y.~Li, D.~Wang, and P.~Qi.
\newblock Bad actor, good advisor: Exploring the role of large language models in fake news detection.
\newblock \emph{arXiv preprint arXiv:2309.12247}, 2023.

\bibitem[Hutto and Gilbert(2014)]{hutto2014vader1}
C.~Hutto and E.~Gilbert.
\newblock Vader: A parsimonious rule-based model for sentiment analysis of social media text.
\newblock In \emph{Proceedings of the international AAAI conference on web and social media}, volume~8, pages 216--225, 2014.

\bibitem[Langguth et~al.(2023)Langguth, Schroeder, Filkukov{\'a}, Brenner, Phillips, and Pogorelov]{langguth2023coco2}
J.~Langguth, D.~T. Schroeder, P.~Filkukov{\'a}, S.~Brenner, J.~Phillips, and K.~Pogorelov.
\newblock Coco: an annotated twitter dataset of covid-19 conspiracy theories.
\newblock \emph{Journal of Computational Social Science}, pages 1--42, 2023.

\bibitem[Lei et~al.(2023)Lei, Dong, Wang, Wang, and Wang]{lei2023instructerc4}
S.~Lei, G.~Dong, X.~Wang, K.~Wang, and S.~Wang.
\newblock Instructerc: Reforming emotion recognition in conversation with a retrieval multi-task llms framework.
\newblock \emph{arXiv preprint arXiv:2309.11911}, 2023.

\bibitem[Liu et~al.(2023)Liu, Zhang, and Liu]{liu2023emotion6}
F.~Liu, X.~Zhang, and Q.~Liu.
\newblock An emotion-aware approach for fake news detection.
\newblock \emph{IEEE Transactions on Computational Social Systems}, 2023.

\bibitem[Liu et~al.(2019)Liu, Ott, Goyal, Du, Joshi, Chen, Levy, Lewis, Zettlemoyer, and Stoyanov]{liu2019roberta7}
Y.~Liu, M.~Ott, N.~Goyal, J.~Du, M.~Joshi, D.~Chen, O.~Levy, M.~Lewis, L.~Zettlemoyer, and V.~Stoyanov.
\newblock Roberta: A robustly optimized bert pretraining approach.
\newblock \emph{arXiv preprint arXiv:1907.11692}, 2019.

\bibitem[Liu et~al.(2024{\natexlab{a}})Liu, Yang, Zhang, Xie, Yu, and Ananiadou]{liu2024emollms}
Z.~Liu, K.~Yang, T.~Zhang, Q.~Xie, Z.~Yu, and S.~Ananiadou.
\newblock Emollms: A series of emotional large language models and annotation tools for comprehensive affective analysis.
\newblock \emph{arXiv preprint arXiv:2401.08508}, 2024{\natexlab{a}}.

\bibitem[Liu et~al.(2024{\natexlab{b}})Liu, Zhang, Yang, Thompson, Yu, and Ananiadou]{liu2024emotion3}
Z.~Liu, T.~Zhang, K.~Yang, P.~Thompson, Z.~Yu, and S.~Ananiadou.
\newblock Emotion detection for misinformation: A review.
\newblock \emph{Information Fusion}, page 102300, 2024{\natexlab{b}}.

\bibitem[Loshchilov and Hutter(2017)]{loshchilov2017decoupled4}
I.~Loshchilov and F.~Hutter.
\newblock Decoupled weight decay regularization.
\newblock \emph{arXiv preprint arXiv:1711.05101}, 2017.

\bibitem[Miani et~al.(2021)Miani, Hills, and Bangerter]{miani2021loco3}
A.~Miani, T.~Hills, and A.~Bangerter.
\newblock Loco: The 88-million-word language of conspiracy corpus.
\newblock \emph{Behavior research methods}, pages 1--24, 2021.

\bibitem[Mompelat et~al.(2022)Mompelat, Tian, Kessler, Luettgen, Rajanala, K{\"u}bler, and Seelig]{mompelat2022loco4}
L.~Mompelat, Z.~Tian, A.~Kessler, M.~Luettgen, A.~Rajanala, S.~K{\"u}bler, and M.~Seelig.
\newblock How “loco” is the loco corpus? annotating the language of conspiracy theories.
\newblock In \emph{Proceedings of the 16th Lingusitic Annotation Workshop (LAW-XVI) within LREC2022}, pages 111--119, 2022.

\bibitem[M{\"u}ller et~al.(2023)M{\"u}ller, Salath{\'e}, and Kummervold]{muller2023covid1}
M.~M{\"u}ller, M.~Salath{\'e}, and P.~E. Kummervold.
\newblock Covid-twitter-bert: A natural language processing model to analyse covid-19 content on twitter.
\newblock \emph{Frontiers in Artificial Intelligence}, 6:\penalty0 1023281, 2023.

\bibitem[Napolitano and Reuter(2023)]{napolitano2023conspiracy1}
M.~G. Napolitano and K.~Reuter.
\newblock What is a conspiracy theory?
\newblock \emph{Erkenntnis}, 88\penalty0 (5):\penalty0 2035--2062, 2023.

\bibitem[Pavlyshenko(2023)]{pavlyshenko2023analysis1}
B.~M. Pavlyshenko.
\newblock Analysis of disinformation and fake news detection using fine-tuned large language model.
\newblock \emph{arXiv preprint arXiv:2309.04704}, 2023.

\bibitem[Penedo et~al.(2023)Penedo, Malartic, Hesslow, Cojocaru, Cappelli, Alobeidli, Pannier, Almazrouei, and Launay]{penedo2023refinedweb}
G.~Penedo, Q.~Malartic, D.~Hesslow, R.~Cojocaru, A.~Cappelli, H.~Alobeidli, B.~Pannier, E.~Almazrouei, and J.~Launay.
\newblock The refinedweb dataset for falcon llm: outperforming curated corpora with web data, and web data only.
\newblock \emph{arXiv preprint arXiv:2306.01116}, 2023.

\bibitem[Peskine et~al.(2021)Peskine, Alfarano, Harrando, Papotti, and Troncy]{peskine2021detecting5}
Y.~Peskine, G.~Alfarano, I.~Harrando, P.~Papotti, and R.~Troncy.
\newblock Detecting covid-19-related conspiracy theories in tweets.
\newblock \emph{MediaEval}, 2021.

\bibitem[Peskine et~al.(2023)Peskine, Koren{\v{c}}i{\'c}, Grubisic, Papotti, Troncy, and Rosso]{peskine2023definitions8}
Y.~Peskine, D.~Koren{\v{c}}i{\'c}, I.~Grubisic, P.~Papotti, R.~Troncy, and P.~Rosso.
\newblock Definitions matter: Guiding gpt for multi-label classification.
\newblock In \emph{Findings of the Association for Computational Linguistics: EMNLP 2023}, pages 4054--4063, 2023.

\bibitem[Pogorelov et~al.(2021)Pogorelov, Schroeder, Brenner, and Langguth]{pogorelov2021fakenews1}
K.~Pogorelov, D.~T. Schroeder, S.~Brenner, and J.~Langguth.
\newblock Fakenews: Corona virus and conspiracies multimedia analysis task at mediaeval 2021.
\newblock In \emph{Multimedia Benchmark Workshop}, volume~67, 2021.

\bibitem[Rasley et~al.(2020)Rasley, Rajbhandari, Ruwase, and He]{rasley2020deepspeed1}
J.~Rasley, S.~Rajbhandari, O.~Ruwase, and Y.~He.
\newblock Deepspeed: System optimizations enable training deep learning models with over 100 billion parameters.
\newblock In \emph{Proceedings of the 26th ACM SIGKDD International Conference on Knowledge Discovery \& Data Mining}, pages 3505--3506, 2020.

\bibitem[Tan et~al.(2022)Tan, Lee, Anbananthen, and Lim]{tan2022roberta2}
K.~L. Tan, C.~P. Lee, K.~S.~M. Anbananthen, and K.~M. Lim.
\newblock Roberta-lstm: a hybrid model for sentiment analysis with transformer and recurrent neural network.
\newblock \emph{IEEE Access}, 10:\penalty0 21517--21525, 2022.

\bibitem[Touvron et~al.(2023)Touvron, Martin, Stone, Albert, Almahairi, Babaei, Bashlykov, Batra, Bhargava, Bhosale, et~al.]{touvron2023llama2}
H.~Touvron, L.~Martin, K.~Stone, P.~Albert, A.~Almahairi, Y.~Babaei, N.~Bashlykov, S.~Batra, P.~Bhargava, S.~Bhosale, et~al.
\newblock Llama 2: Open foundation and fine-tuned chat models.
\newblock \emph{arXiv preprint arXiv:2307.09288}, 2023.

\bibitem[Workshop et~al.(2022)Workshop, Scao, Fan, Akiki, Pavlick, Ili{\'c}, Hesslow, Castagn{\'e}, Luccioni, Yvon, et~al.]{workshop2022bloom7}
B.~Workshop, T.~L. Scao, A.~Fan, C.~Akiki, E.~Pavlick, S.~Ili{\'c}, D.~Hesslow, R.~Castagn{\'e}, A.~S. Luccioni, F.~Yvon, et~al.
\newblock Bloom: A 176b-parameter open-access multilingual language model.
\newblock \emph{arXiv preprint arXiv:2211.05100}, 2022.

\bibitem[Xie et~al.(2023)Xie, Han, Zhang, Lai, Peng, Lopez-Lira, and Huang]{xie2023pixiu2}
Q.~Xie, W.~Han, X.~Zhang, Y.~Lai, M.~Peng, A.~Lopez-Lira, and J.~Huang.
\newblock Pixiu: A large language model, instruction data and evaluation benchmark for finance.
\newblock \emph{arXiv preprint arXiv:2306.05443}, 2023.

\bibitem[Yanagi et~al.(2021)Yanagi, Orihara, Tahara, Sei, and Ohsuga]{yanagi2021classifying4}
Y.~Yanagi, R.~Orihara, Y.~Tahara, Y.~Sei, and A.~Ohsuga.
\newblock Classifying covid-19 conspiracy tweets with word embedding and bert.
\newblock In \emph{Working Notes Proceedings of the MediaEval 2021 Workshop, Online}, pages 13--15, 2021.

\bibitem[Yang et~al.(2023)Yang, Zhang, Kuang, Xie, and Ananiadou]{yang2023mentalllama3}
K.~Yang, T.~Zhang, Z.~Kuang, Q.~Xie, and S.~Ananiadou.
\newblock Mentalllama: Interpretable mental health analysis on social media with large language models.
\newblock \emph{arXiv preprint arXiv:2309.13567}, 2023.

\bibitem[Yuan et~al.(2023)Yuan, Xie, Huang, and Ananiadou]{yuan2023back1}
C.~Yuan, Q.~Xie, J.~Huang, and S.~Ananiadou.
\newblock Back to the future: Towards explainable temporal reasoning with large language models.
\newblock \emph{arXiv preprint arXiv:2310.01074}, 2023.

\bibitem[Zaeem et~al.(2020)Zaeem, Li, and Barber]{zaeem2020sentiment5}
R.~N. Zaeem, C.~Li, and K.~S. Barber.
\newblock On sentiment of online fake news.
\newblock In \emph{2020 IEEE/ACM International Conference on Advances in Social Networks Analysis and Mining (ASONAM)}, pages 760--767. IEEE, 2020.

\bibitem[Zhang et~al.(2023)Zhang, Yang, Zhou, Ali~Babar, and Liu]{zhang2023enhancing8}
B.~Zhang, H.~Yang, T.~Zhou, M.~Ali~Babar, and X.-Y. Liu.
\newblock Enhancing financial sentiment analysis via retrieval augmented large language models.
\newblock In \emph{Proceedings of the Fourth ACM International Conference on AI in Finance}, pages 349--356, 2023.

\bibitem[Zhang et~al.(2022)Zhang, Roller, Goyal, Artetxe, Chen, Chen, Dewan, Diab, Li, Lin, et~al.]{zhang2022opt6}
S.~Zhang, S.~Roller, N.~Goyal, M.~Artetxe, M.~Chen, S.~Chen, C.~Dewan, M.~Diab, X.~Li, X.~V. Lin, et~al.
\newblock Opt: Open pre-trained transformer language models.
\newblock \emph{arXiv preprint arXiv:2205.01068}, 2022.

\bibitem[Zhang et~al.(2021)Zhang, Cao, Li, Sheng, Zhong, and Shu]{zhang2021mining7}
X.~Zhang, J.~Cao, X.~Li, Q.~Sheng, L.~Zhong, and K.~Shu.
\newblock Mining dual emotion for fake news detection.
\newblock In \emph{Proceedings of the web conference 2021}, pages 3465--3476, 2021.

\bibitem[Zheng et~al.(2023)Zheng, Liao, Deng, and Nie]{zheng2023building4}
Z.~Zheng, L.~Liao, Y.~Deng, and L.~Nie.
\newblock Building emotional support chatbots in the era of llms.
\newblock \emph{arXiv preprint arXiv:2308.11584}, 2023.

\end{thebibliography}


\section{Supplementary Material}

\begin{figure}[htb]
\centering
\includegraphics[width=\columnwidth]{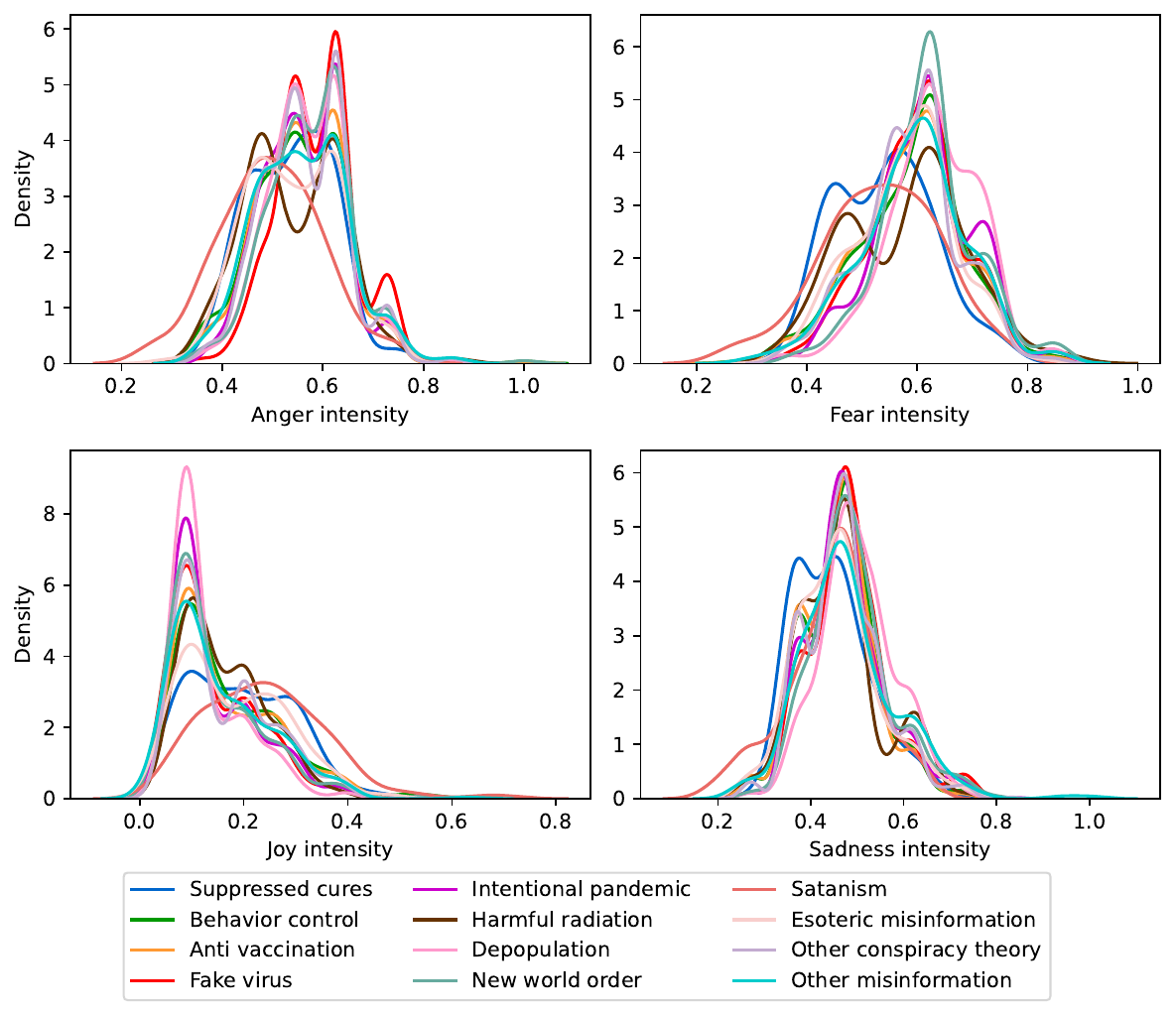}
\caption{Emotion intensity of different COCO conspiracies}
\label{fig:COCO_EIreg_conspiracy}
\end{figure}

\begin{figure}[htb]
\centering
\includegraphics[width=\columnwidth]{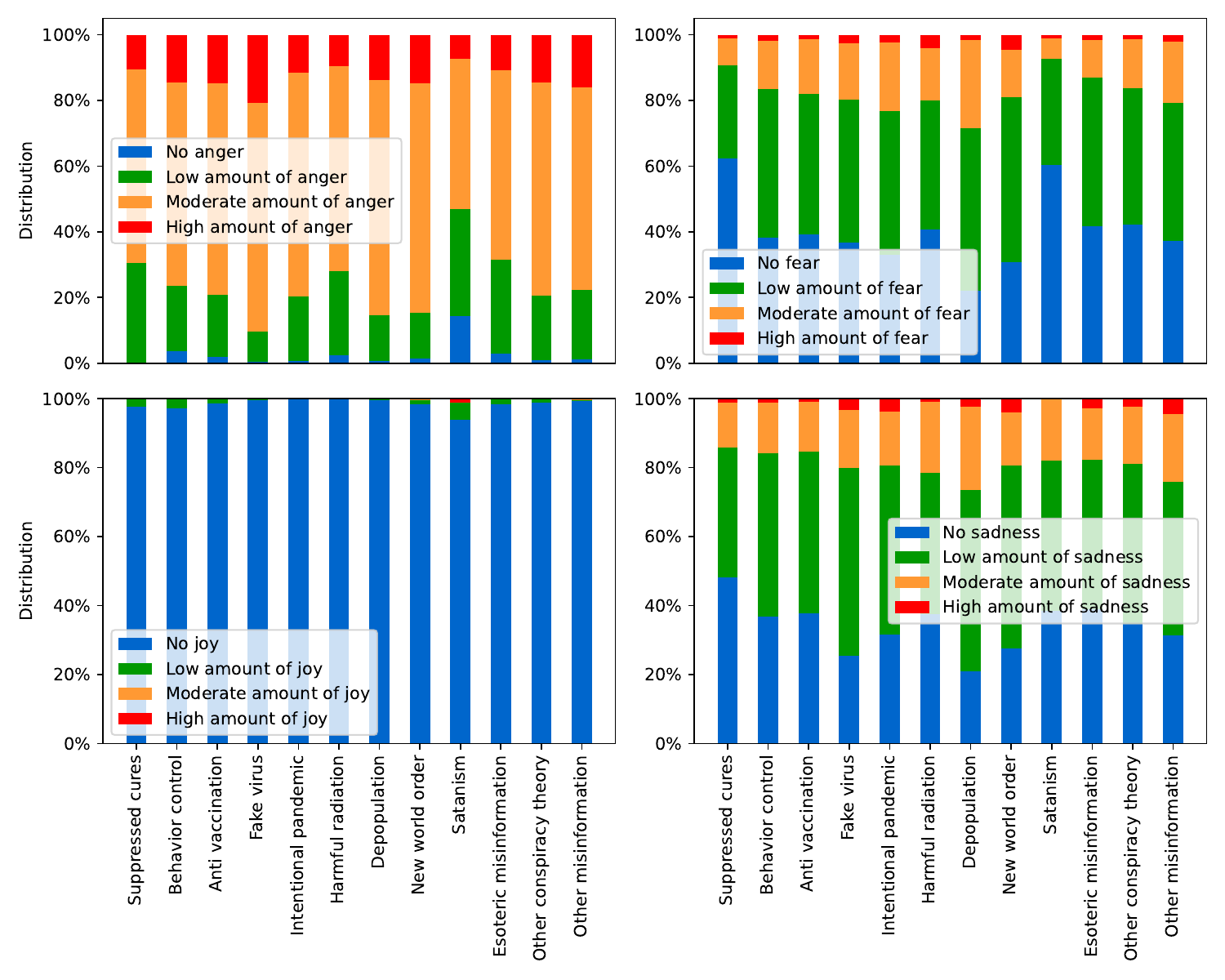}
\caption{Emotion intensity classification of different COCO conspiracies}
\label{fig:COCO_EIoc_conspiracy}
\end{figure}

\begin{figure}[htb]
\centering
\includegraphics[width=\columnwidth]{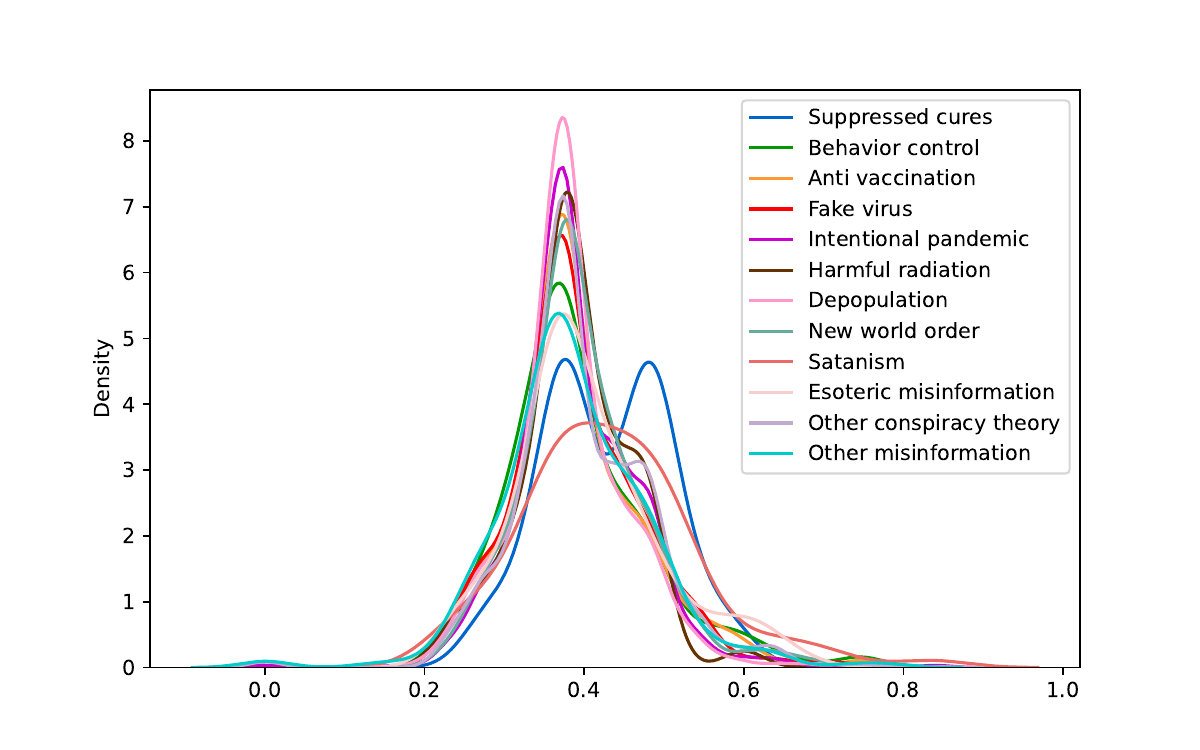}
\caption{Sentiment strength of different COCO conspiracies}
\label{fig:COCO_Vreg_conspiracy}
\end{figure}

\begin{figure}[!t]
\centering
\includegraphics[width=\columnwidth]{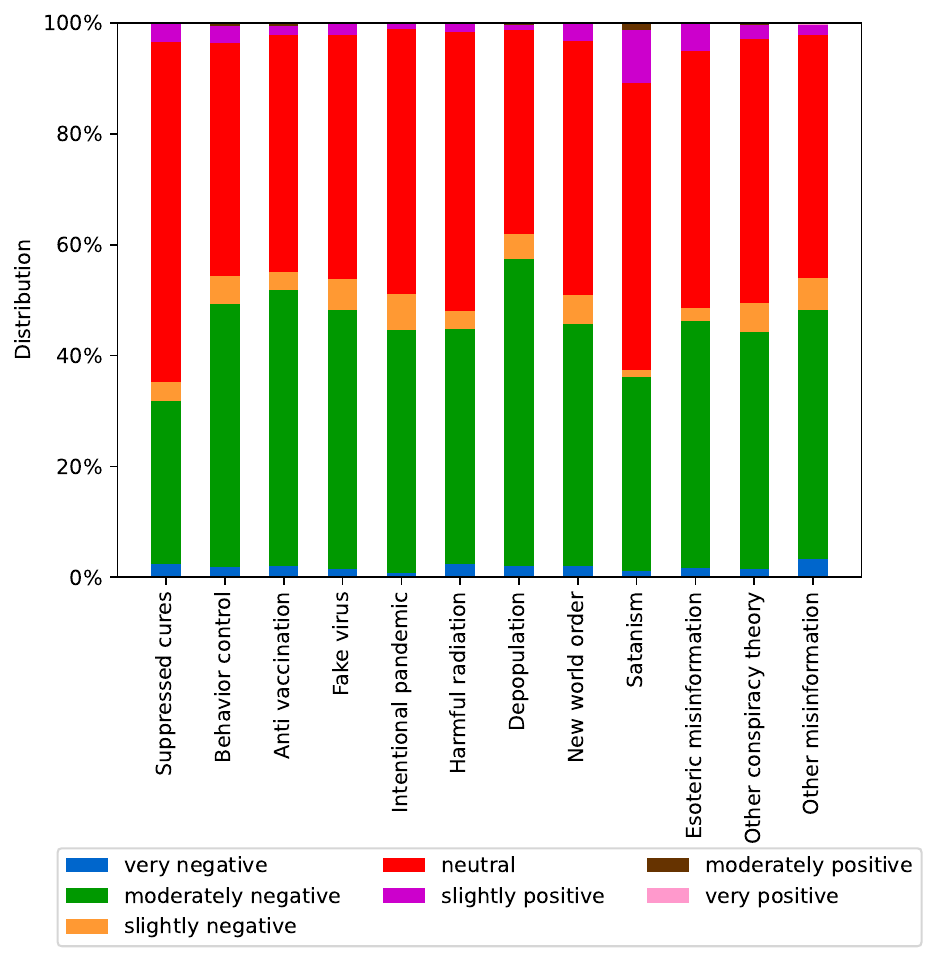}
\caption{Sentiment classification of different COCO conspiracies}
\label{fig:COCO_Voc_conspiracy}
\end{figure}

\begin{figure}[!t]
\centering
\includegraphics[width=\columnwidth]{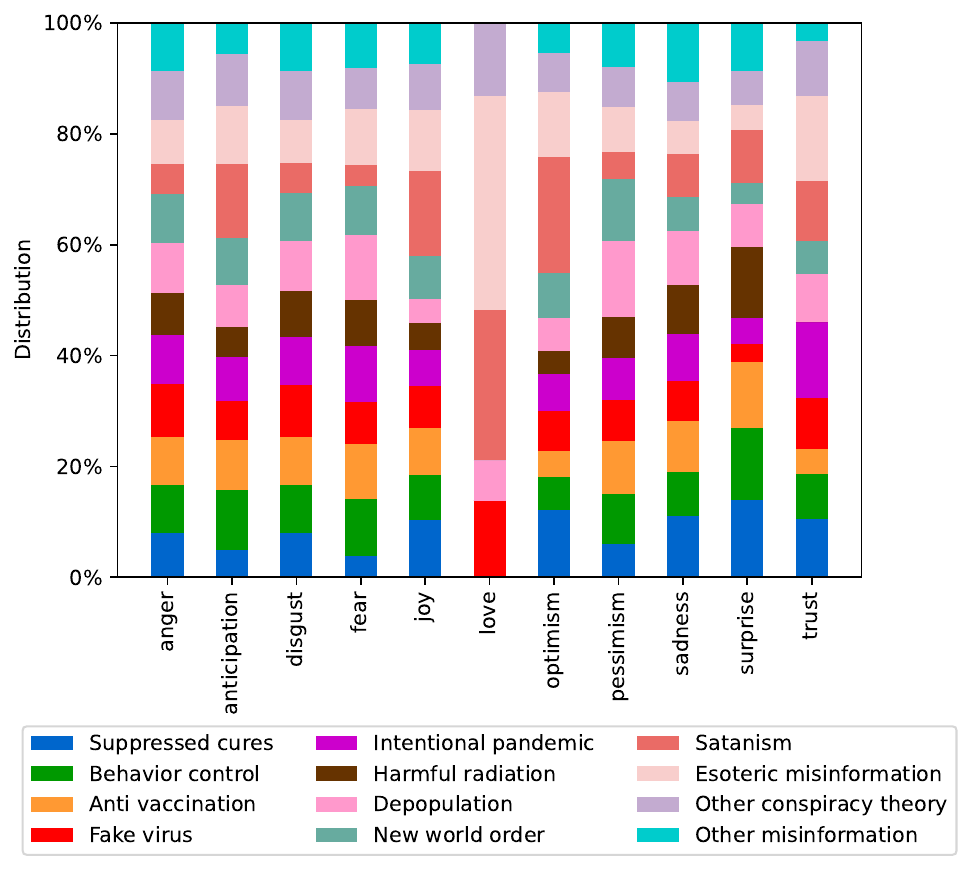}
\caption{Emotion classification of different COCO conspiracies}
\label{fig:COCO_Ec_conspiracy}
\end{figure}



\begin{figure}[!t]
\centering
\includegraphics[width=\columnwidth]{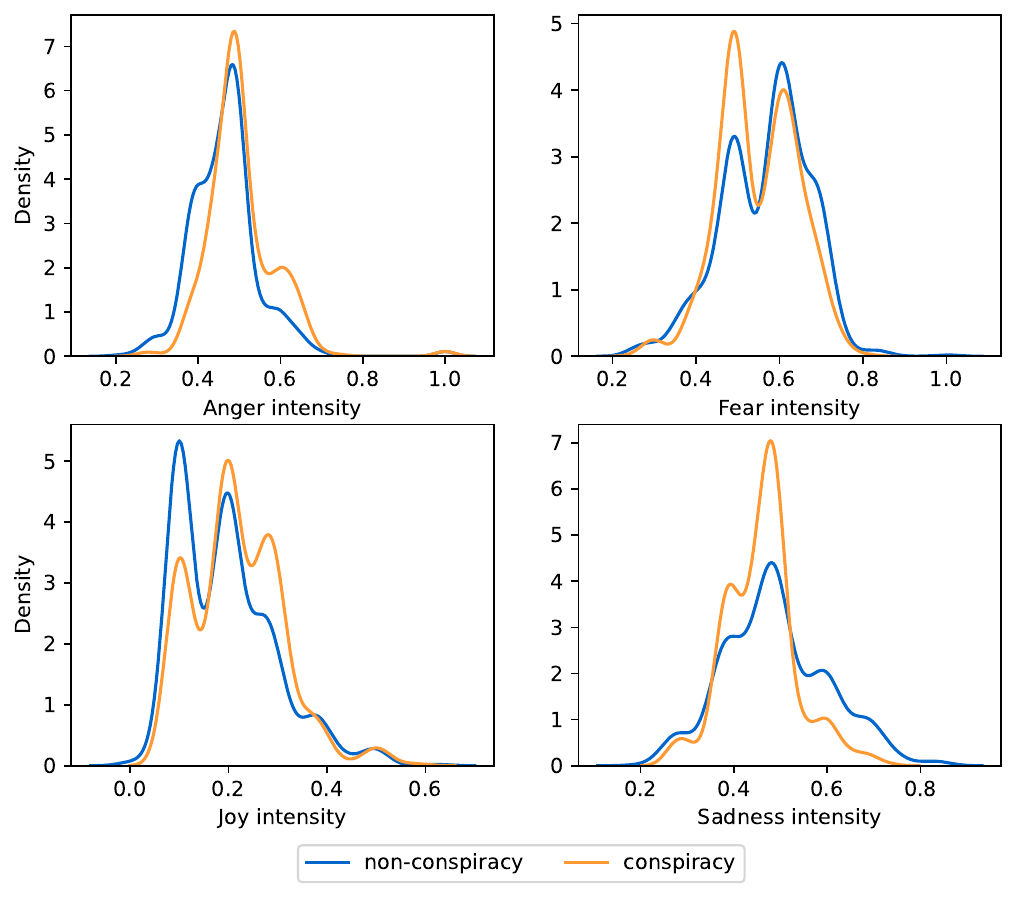}
\caption{Emotion intensity of conspiracy and non-conspiracy in LOCOAnnotation}
\label{fig:LOCO_EI_label}
\end{figure}

\begin{figure}[!t]
\centering
\includegraphics[width=\columnwidth]{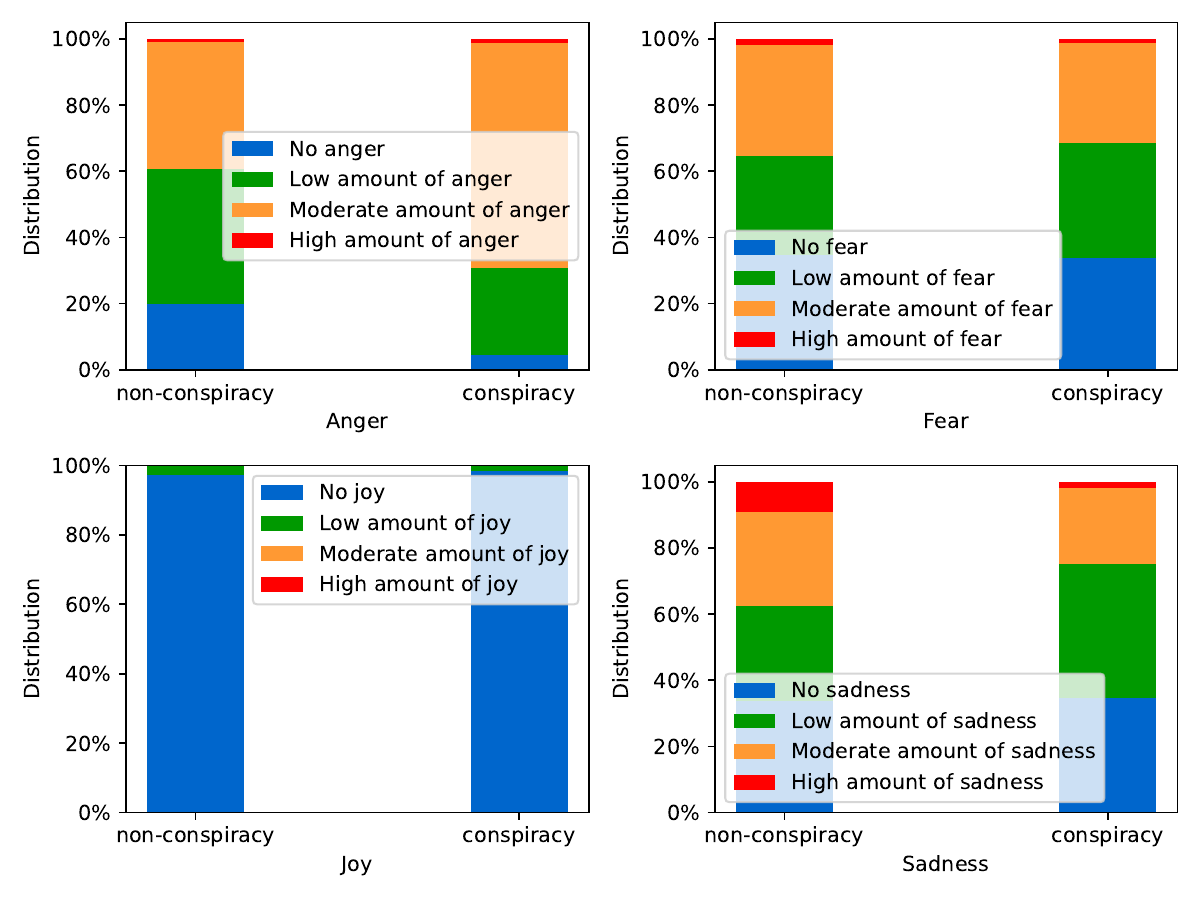}
\caption{Emotion intensity classification of conspiracy and non-conspiracy  in LOCOAnnotation}
\label{fig:LOCO_EIoc_label}
\end{figure}

\begin{figure}[!t]
\centering
\includegraphics[width=\columnwidth]{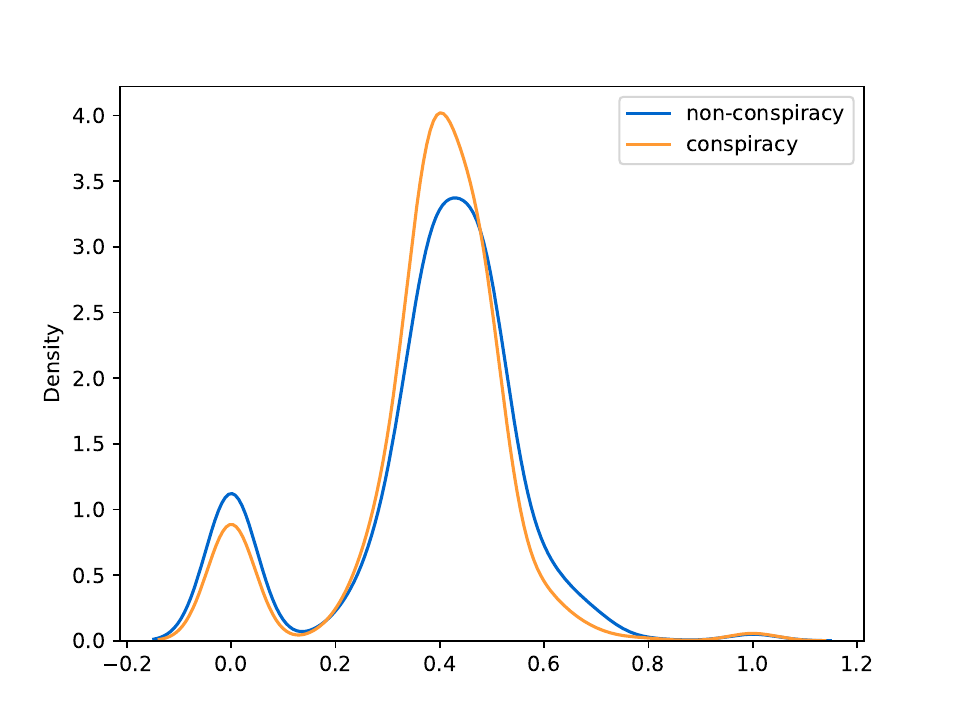}
\caption{Sentiment strength of conspiracy and non-conspiracy in LOCOAnnotation}
\label{fig:LOCO_Vreg_label}
\end{figure}

\begin{figure}[!t]
\centering
\includegraphics[width=\columnwidth]{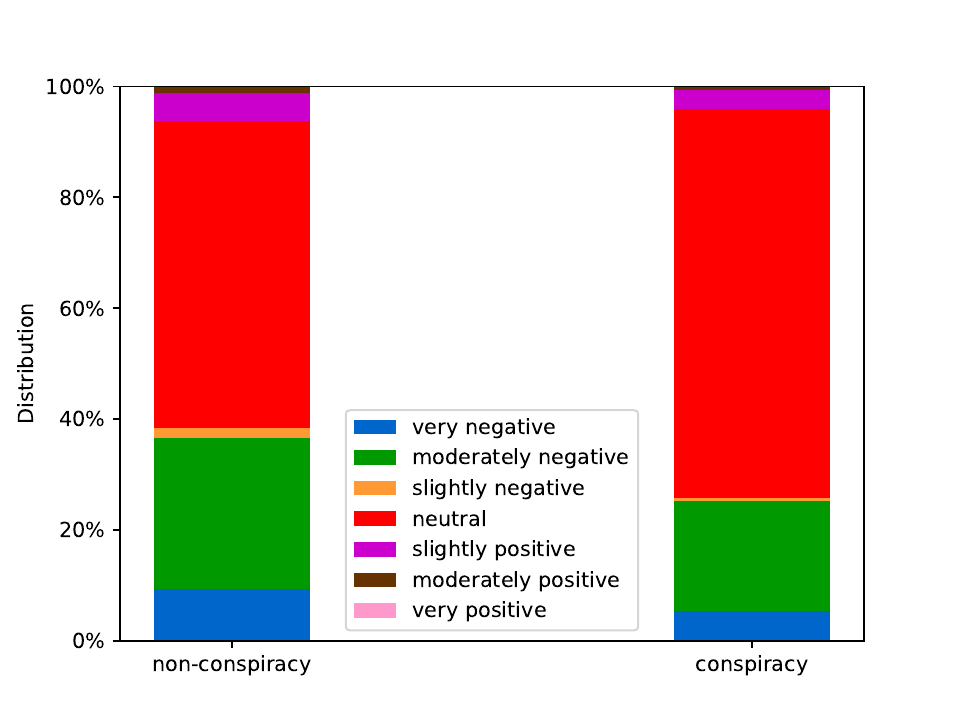}
\caption{Sentiment classification of conspiracy and non-conspiracy in LOCOAnnotation}
\label{fig:LOCO_Voc_label}
\end{figure}

\begin{figure}[!t]
\centering
\includegraphics[width=\columnwidth]{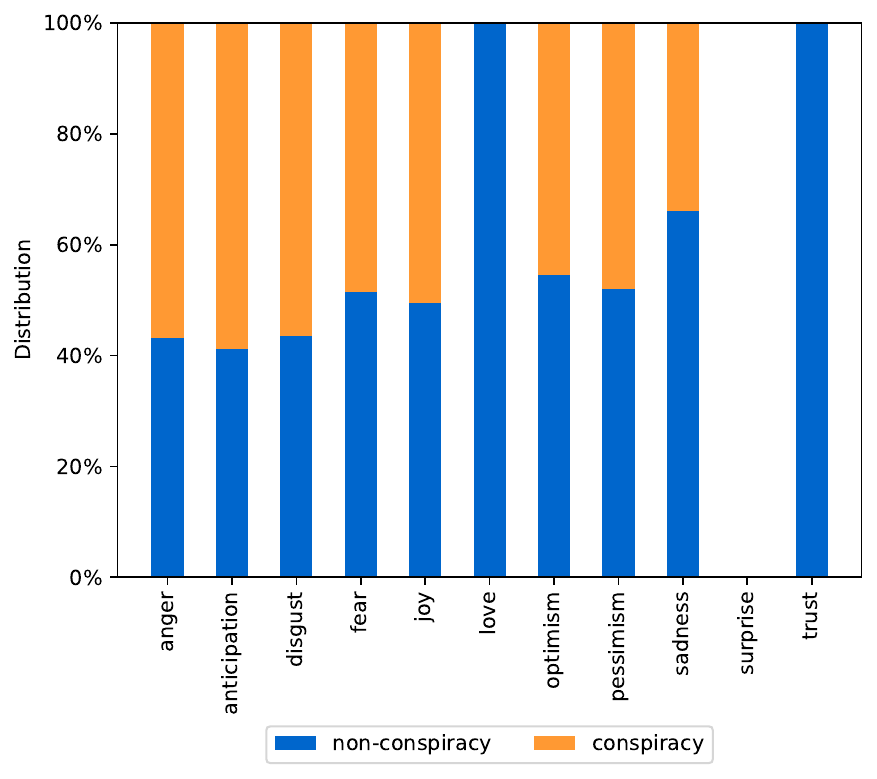}
\caption{Emotion classification of conspiracy and non-conspiracy in LOCOAnnotation}
\label{fig:LOCO_Ec_label}
\end{figure}


\begin{figure}[!t]
\centering
\includegraphics[width=\columnwidth]{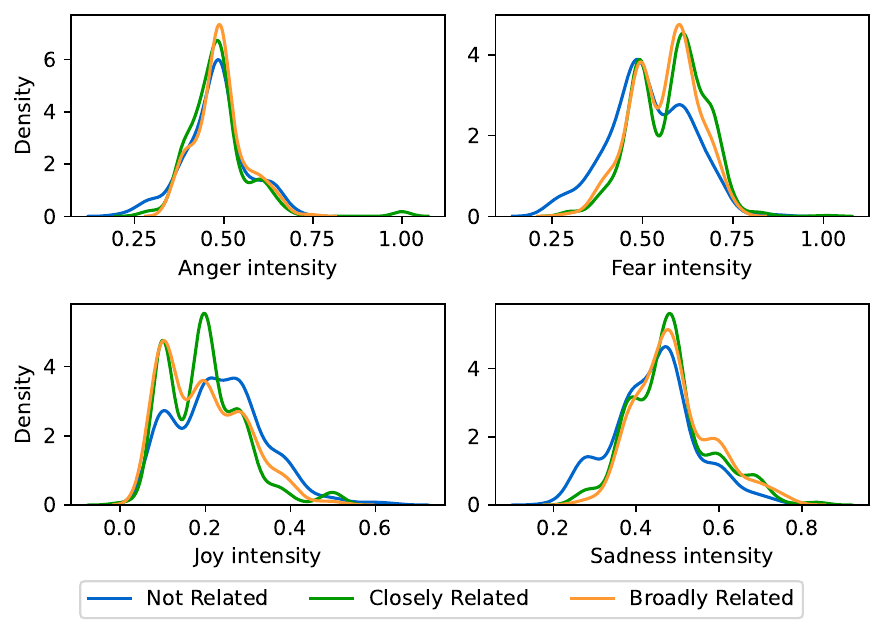}
\caption{Emotion intensity of different relatedness in LOCOAnnotation}
\label{fig:LOCO_EIreg_relatedness}
\end{figure}

\begin{figure}[!t]
\centering
\includegraphics[width=\columnwidth]{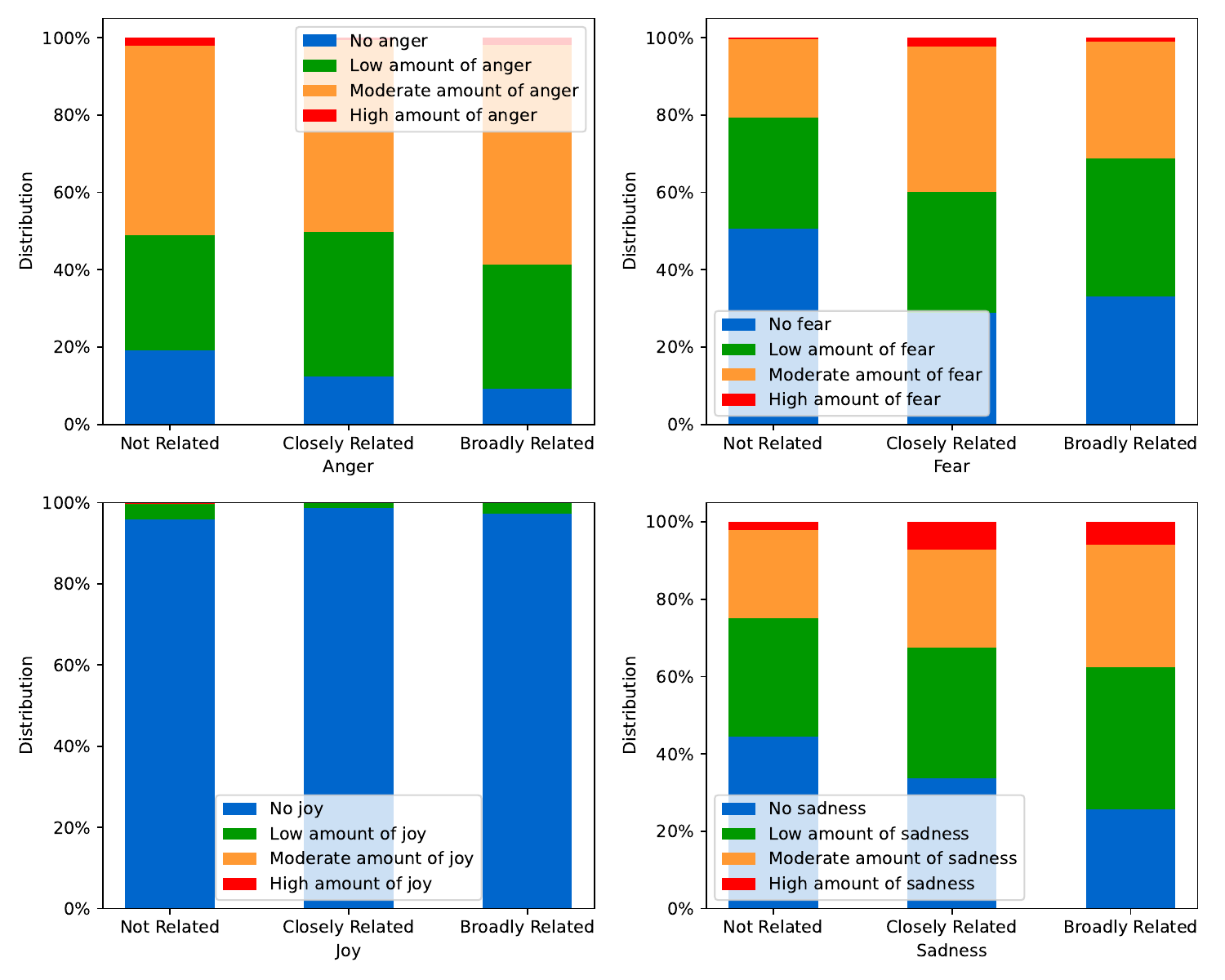}
\caption{Emotion intensity classification of different relatedness in LOCOAnnotation}
\label{fig:LOCO_EIoc_relatedness}
\end{figure}

\begin{figure}[!t]
\centering
\includegraphics[width=\columnwidth]{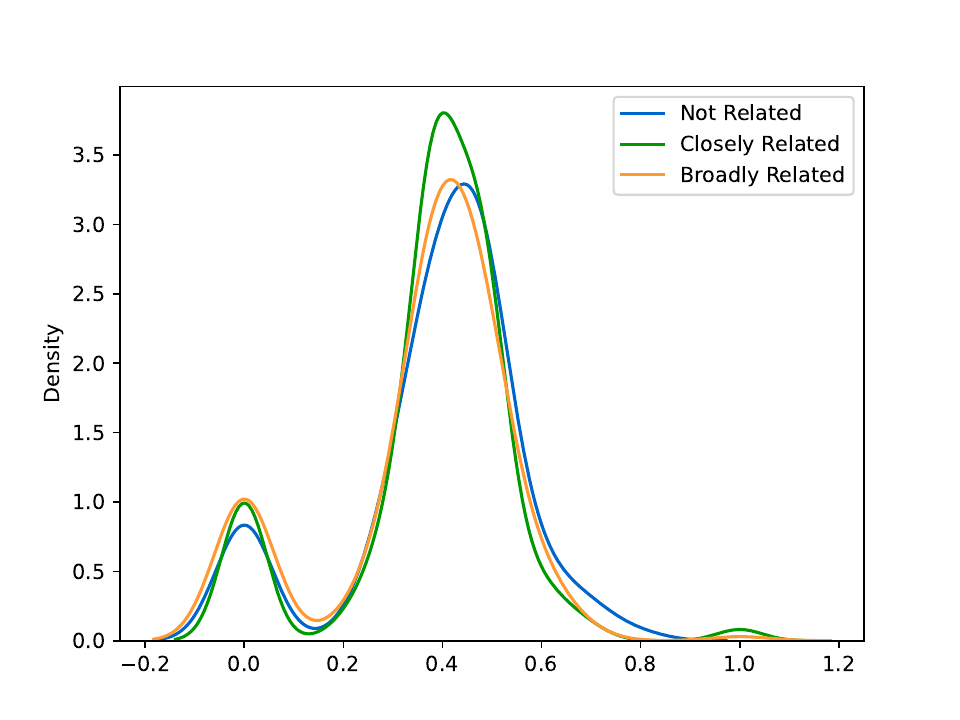}
\caption{Sentiment strength of different relatedness in LOCOAnnotation}
\label{fig:LOCO_Vreg_relatedness}
\end{figure}

\begin{figure}[!t]
\centering
\includegraphics[width=\columnwidth]{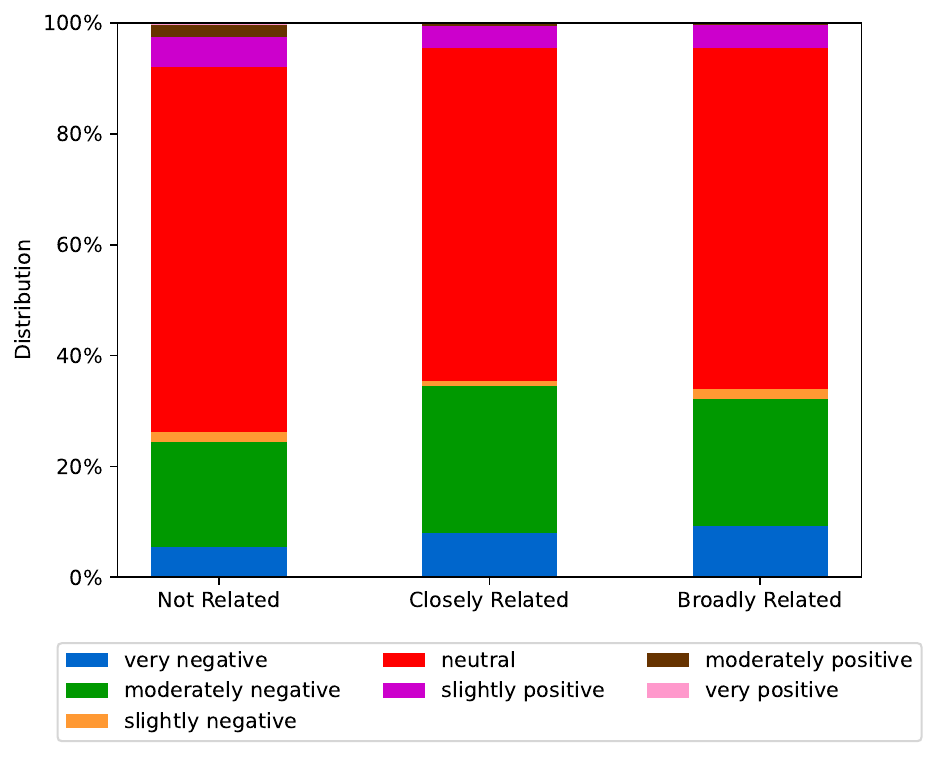}
\caption{Sentiment classification of different relatedness in LOCOAnnotation}
\label{fig:LOCO_Voc_relatedness}
\end{figure}

\begin{figure}[!t]
\centering
\includegraphics[width=\columnwidth]{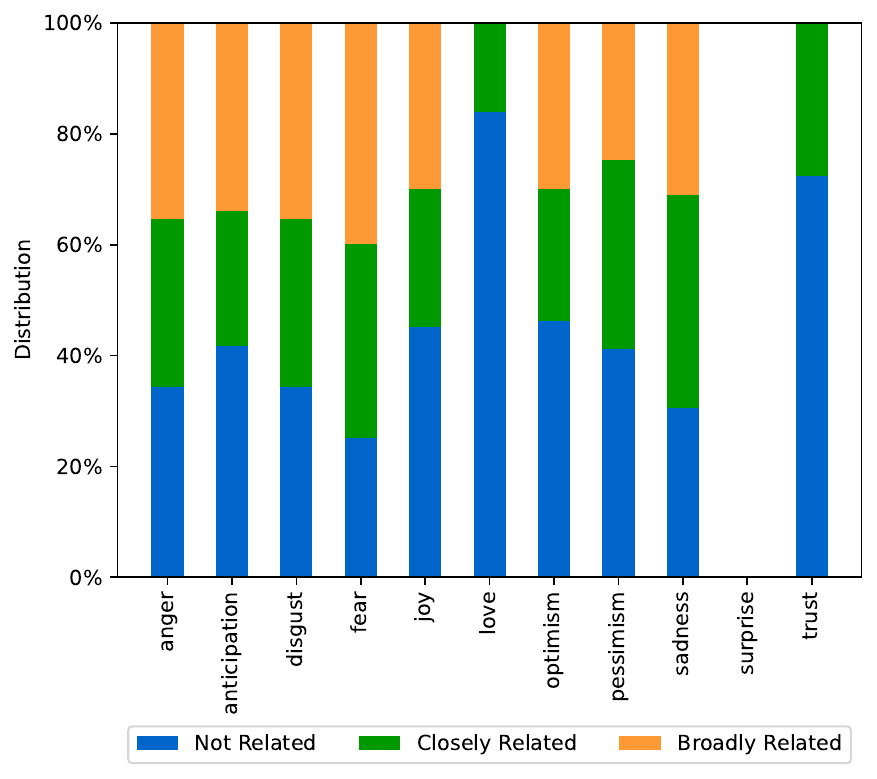}
\caption{Emotion classification of different relatedness in LOCOAnnotation}
\label{fig:LOCO_Ec_relatedness}
\end{figure}

\end{document}